\documentclass[10pt]{article}  

\newif\ifarXiv         
\newif\ifjournal        

  \journalfalse \arXivtrue  

\ifjournal 
\input{macro_journal}
\fi

\ifarXiv     
\fi

\usepackage{amssymb,amsthm}    
\usepackage{graphicx}          
\usepackage{amsmath}         
\usepackage{caption}
\usepackage{subcaption}
\usepackage{ dsfont }
\usepackage{listings}
\usepackage{titlepic}
\usepackage{xcolor}
\usepackage{algorithm}
\usepackage{algorithmicx}
\usepackage{todonotes}
\usepackage{array}
\usepackage{booktabs}
\usepackage{bm,tabularx,cases}

\numberwithin{equation}{section}

\allowdisplaybreaks[3]

\DeclareMathOperator{\assump}{(A1)}
\DeclareMathOperator{\assumpp}{(A2)}
\DeclareMathOperator{\assumppp}{(A3)}
\usepackage{hyperref}
\hypersetup{
    colorlinks=true,
    linkcolor=blue,
    filecolor=magenta,      
    urlcolor=cyan,
    citecolor = blue,
}

\usepackage{ dsfont }
\usepackage{listings}
\usepackage{bbm}
\usepackage{color}
\usepackage{ dsfont }
\usepackage{enumerate}
\usepackage{todonotes}
\DeclareMathOperator*{\argmin}{arg\,min}

\usepackage{booktabs}  

\renewcommand{\Delta}{\triangle}

\def\data{{\scriptscriptstyle \mathcal{D} }}

\definecolor{darkblue}{rgb}{0,0,0.7}

\definecolor{darkgreen}{rgb}{0.01,0.75,0.24}

\def \Ee[#1]{\mathcal{E}^{\text{{#1}}}}
\def\R{\mathbf{R}}

\def\pa[#1,#2]{\frac{\partial {#1}}{\partial {#2}} }

\def\idom[#1,#2,#3]{\int_{#1}\hspace{1pt} {#2} \hspace{1pt} \text{d}{#3}}
\def\res[#1,#2]{\left.{#1}\right|_{#2}}

\def\var[#1,#2]{\langle \delta \mathcal{E}^{\text{{#1}}}({#2}),v\rangle}
\def\vars[#1,#2,#3]{\langle \delta^2\mathcal{E}^{\text{{#1}}}({#2})v,{#3}\rangle}
\def\vard[#1,#2,#3,#4]{\langle \delta\mathcal{E}^{\text{{#1}}}({#2})-\delta\mathcal{E}^{\text{{#3}}}({#4}),v\rangle}

\def\E{\mathbb{E}}

\def\mH{\mathcal{H}}

\def\mS{\mathcal{S}}

\def\mN{\mathcal{N}}

\newcommand{\be}{\begin{equation}}
\newcommand{\en}{\end{equation}}
\newcommand{\ben}{\begin{equation*}}
\newcommand{\enn}{\end{equation*}}
\newcommand{\bea}{\begin{aligned}}
\newcommand{\ena}{\end{aligned}}

\def\ba#1\ena{\begin{align}#1\end{align}}
\def\ban#1\enan{\begin{align*}#1\end{align*}}

\newtheorem{theorem}{Theorem}[section]
\newtheorem{definition}[theorem]{Definition}   

\newtheorem{lemma}[theorem]{Lemma}
\newtheorem{myassumption}[theorem]{Assumption}
\newtheorem{proposition}[theorem]{Proposition}
\newtheorem{remark}[theorem]{Remark}
\newtheorem{example}[theorem]{Example}

\numberwithin{equation}{section}

\def\calE{\mathcal{E}} 

\def\calN{\mathcal{N}}
\def\calQ{\mathcal{Q}}

\newcommand{\norm}[1]{\left\|#1\right\|}
\newcommand{\abs}[1]{\left|#1\right|}

\newcommand{\innerp}[1]{\langle{#1}\rangle}
\newcommand{\dbinnerp}[1]{\langle\hspace{-1mm}\langle{#1}\rangle \hspace{-1mm}\rangle}

\def\calS{\mathcal{S}}

\newcommand{\mbf}[1]{\boldsymbol{#1}}
\def\Gbar{ {\overline{G}} }

\def\Abar{ {\overline{A}} }
\def\bbar{ {\overline{b}} }
\def\LGbar{ {\mathcal{L}_{\overline{G}}}  }

\def\mH{\mathcal{H}}

\def\R{\mathbb{R}}

\def\spaceX{\mathbb{X}}
\def\spaceY{\mathbb{Y}}

\def\prior{\pi_0}
\def\posterior{\pi_1}
\def\likelihood{\pi_L}

\title{A Data-Adaptive Prior for Bayesian Learning \\ of Kernels in Operators}
\bigskip
\author{Neil K. Chada\thanks{Department of Actuarial Mathematics and Statistics, Heriot Watt University, Edinburgh, EH14 4AS, UK  (\texttt{\textcolor{black}{neilchada123@gmail.com}})} 
 \and  Quanjun Lang\thanks{Department of Mathematics, Johns Hopkins University, Baltimore, MD 21218, USA
 (\texttt{\textcolor{black}{qlang1@jhu.edu}}), (\texttt{\textcolor{black}{feilu@math.jhu.edu}}), (\texttt{\textcolor{black}{xiongwang@jhu.edu}})}
  \and Fei Lu\footnotemark[2]
    \and Xiong Wang\footnotemark[2]}

\usepackage{enumitem}

\begin{document}
\maketitle


\begin{abstract}
Kernels effectively represent nonlocal dependencies and are extensively employed in formulating operators between function spaces. Thus, learning kernels in operators from data is an inverse problem of general interest. Due to the nonlocal dependence, the inverse problem is often severely ill-posed with a data-dependent normal operator. Traditional Bayesian methods address the ill-posedness by a non-degenerate prior, which may result in an unstable posterior mean in the small noise regime, especially when data induces a perturbation in the null space of the normal operator. We propose a new data-adaptive Reproducing Kernel Hilbert Space (RKHS) prior, which ensures the stability of the posterior mean in the small noise regime. We analyze this adaptive prior and showcase its efficacy through applications on Toeplitz matrices and integral operators. Numerical experiments reveal that fixed non-degenerate priors can produce divergent posterior means under errors from discretization, model inaccuracies, partial observations, or erroneous noise assumptions. In contrast, our data-adaptive RKHS prior consistently yields convergent posterior means.

  \end{abstract}

\textbf{Keywords}: Data-adaptive prior,  kernels in operators, linear Bayesian inverse problem, \\ RKHS, Tikhonov regularization \\\ 
\textbf{2020 Mathematics Subject Classification}: 62F15, 47A52, 47B32

 \vspace{-1mm}
 \tableofcontents

\section{Introduction}\label{sec:intro}

Kernels are efficient in representing nonlocal or long-range dependence and interaction between high- or infinite-dimensional variables. Thus, they are widely used to design operators between function spaces, with numerous applications in machine learning such as kernel methods (e.g., \cite{belkin2018understand,CZ07book,darcy2021learning,hofmann2008kernel,Sriperumbudur2011Universality,owhadi2019kernel}) and 
operator learning (e.g., \cite{KLA21,LJP21}), in partial differential equations (PDEs) and stochastic processes such as nonlocal and fractional diffusions (e.g.,  \cite{bucur2016_NonlocalDiffusion,delia2020_NumericalMethods,du2012_AnalysisApproximation,you2022_DatadrivenPeridynamic,you2021_DatadrivenLearning}), and in multi-agent systems (e.g., \cite{carrillo2019aggregation,LMT21_JMLR,LZTM19pnas,MT14}).

Learning kernels in operators from data is an integral part of these applications. We consider the case when the operator depends on the kernel linearly, and the learning is a linear inverse problem. However, the inverse problem is often severely ill-posed, due to the nonlocal dependence and the presence of various perturbations resulting from noise in data,  numerical error, or model error.  To address the ill-posedness, a Bayesian approach or a variational approach with regularization is often used. In either approach, the major challenge is the selection of a prior or a regularization term since there is limited prior knowledge about the kernel.

This study examines the selection of the prior in a Bayesian approach. The common practice is to use a non-degenerate prior. However, we show that a non-degenerate prior may result in an unstable posterior mean in the small noise regime, especially when data induces a perturbation in the null space of the normal operator.  

We propose a new data-adaptive Reproducing Kernel Hilbert Space (RKHS) prior, which ensures the stability of the posterior mean in the small noise regime. We analyze this adaptive prior and showcase its efficacy through applications on learning kernels in Toeplitz matrices and integral operators. 

 \vspace{-2mm}
\subsection{Problem setup}

We aim to learn the kernel $\phi$ in the operator $R_\phi: \spaceX\to \spaceY$ in the following model 
\begin{equation}\label{eq:map_R}
 R_\phi[u]  + \eta  + \xi = f,
 \end{equation}
by fitting the model to the data consisting of $N$ input-output pairs: 
\begin{equation}\label{eq:data}
 \mathcal{D}  = \{(u^{k},f^{k})\}_{k=1}^N, \quad (u^k,f^k)\in \spaceX \times \spaceY. 
\end{equation}
Here  $\spaceX$ is a Banach space,  $\spaceY$ is a Hilbert space, the measurement noise $\eta $ is a $\spaceY$-valued white noise in the sense that $\E[\innerp{\eta, f}_\spaceY^2] = \sigma_\eta^2\innerp{f, f}_\spaceY$ for any $f\in\spaceY$. The term $\xi$ represents unknown model errors such as model misspecification or computational error due to incomplete data, and it may depend on the input data $u$.  

The operator $R_\phi$ depends \emph{non-locally} on the kernel $\phi$ in the form 
\begin{equation}\label{eq:operator_general}
R_\phi[u](y) = \int_{\Omega} \phi(y-x) g[u](x,y)\mu(dx), \quad \forall y\in \Omega,
\end{equation}
where $(\Omega,\mu)$ is a measure space that can be either a domain in the Euclidean space with the Lebesgue measure or a discrete set with an atomic measure. {For simplicity, we let $\mathbb{Y} = L^2(\Omega,\mu)$ throughout this study.} Here $g[u]$ is a bounded (nonlinear) functional of $u$ and is assumed to be known.  Note that the operator $R_\phi$ can be nonlinear in $u$, but it depends linearly on $\phi$. 
Such operators are widely seen in PDEs, matrix operators, and image processing. Examples include the Toeplitz matrix, integral and nonlocal operators; see Sect.\ref{sec:learningkernels}. In these examples, there is often limited prior knowledge about the kernel.       

The inverse problem of learning the kernel $\phi$ is often ill-posed, due to the nonlocal dependence of the output data $f$ on the kernel. The ill-posedness, in terms of minimizing the negative log-likelihood of the data, 
\begin{equation}\label{eq:lossFn0}
 \calE(\phi) = \frac{1}{N\sigma_\eta^2}\sum_{1\leq k\leq N} \|R_\phi[u^k]-f^k\|_{\spaceY}^2=
 \frac{1}{2\sigma_{\eta}^{2}}  \left[ \innerp{\LGbar \phi,\phi}_{L^2_\rho} -2 \innerp{\phi^\data,\phi}_{L^2_\rho}+C_N^f \right], 
\end{equation}
appears as the instability of the minimizer $\LGbar^{-1}\phi^\data$ when it exists. Here $L^2_\rho$ is space of square-integrable functions with measure $\rho$,  the normal operator $\LGbar$ is a trace-class operator, and $\phi^\data$ comes from data; see Sect.\ref{sec:fsoi} for details. Thus, the ill-posedness is rooted in the unboundedness of $\LGbar^{-1}$ and the perturbation of $\phi^\data$. 

There are two common strategies to overcome the ill-posedness: a prior in a Bayesian approach and regularization in a variational approach; see, e.g., \cite{Stuart10,hansen1994_regularization_tools,bauer2007regularization} and a sample of the large literature in Sect.\ref{sec:relatedwork}). 

This study focuses on selecting a prior for the Bayesian approach. The major challenge is the limited prior information about the kernel and the need to overcome the ill-posedness caused by a data-dependent normal operator. 

\subsection{Proposed: a data-adaptive RKHS prior}
Due to the lack of prior information about the kernel, one may use the default prior, a \emph{non-degenerate prior}, e.g., a Gaussian distribution $\calN(0,\calQ_0)$ with a nondegenerate covariance operator $\calQ_0$, with the belief that it is a safe choice to ensure a well-defined posterior.  

However, we show that the fixed non-degenerate prior has the risk of leading to a divergent posterior mean in the small noise limit. Specifically, the posterior mean, 
$$\mu_1 = (\LGbar + \sigma_\eta^2 \calQ_0)^{-1}\phi^\data,$$
obtained from the prior  $\calN(0,\calQ_0)$ and the likelihood that yields \eqref{eq:lossFn0}, 
blows up when, the variance of the noise, $\sigma_\eta\to 0$ if $\phi^\data$ contains a perturbation in the null space of the normal operator $\LGbar$; see Proposition \ref{thm:risk_prior}. Such perturbation can be caused by any of the four types of errors in data or computation: (a) discretization error, (b) model error, (c) partial observations, and (d) wrong noise assumption. Thus, a prior adaptive to $\LGbar$ and $\phi^\data$ is needed to remove the risk. 

We propose a \emph{data-adaptive RKHS prior} that ensures a stable posterior mean in the small noise regime. It is a Gaussian distribution $ \calN(0, \lambda_*^{-1} \LGbar)$ with the parameter $\lambda_*$ selected adaptive to $\phi^\data$. We prove in Theorem \ref{thm:MAP_stability} that it leads to a stable posterior whose mean 
$$\mu_1^\data =( \LGbar^2 + \sigma_\eta^2 \lambda_* I_{\mathrm{Null(\LGbar)}^\perp } )^{-1} \LGbar \phi^\data, $$
always has a small noise limit, and the small noise limit converges to the identifiable parts of the true kernel. Furthermore, we show that our prior outperforms the non-degenerate prior in producing a more accurate posterior mean and smaller posterior uncertainty in terms of the trace of the posterior covariance; see Sect.\ref{sec:quality_posterior}. The prior is called an RKHS prior because its Cameron-Martin space $\LGbar^{1/2} L^2_\rho$ is the RKHS with a reproducing kernel $\Gbar$ determined by the operator $R_\phi$ and the data. Importantly, the closure of this RKHS is the space in which the components of the true kernel can be identified from data. 

We also study the computational practice of the data-adaptive prior and demonstrate it on the Toeplitz matrices and integral operators. We select the hyper-parameter by the L-curve method in \cite{hansen_LcurveIts_a}. Numerical tests show that while a fixed non-degenerate prior leads to divergent posterior means, the data-adaptive prior always attains stable posterior means with small noise limits; see Sect.\ref{sec:num}.

The outline of this study is as follows. Sect.\ref{sec:setting} presents the mathematical setup of this study, and shows the ill-posedness of the inverse problem through the variational approach. We introduce in Sect.\ref{sec:Bayes} the Bayesian approach and show the issue of a fixed non-degenerate prior. To solve the issue, we introduce a data-adaptive prior in Sect.\ref{sec:DAprior}, and analyze its advantages. 
Sect.\ref{sec:num} discusses the computational practice and demonstrates the advantage of the data-adaptive prior in numerical tests on  Toeplitz matrices and integral operators. 
Finally, Sect.\ref{sec:conc} concludes our findings and provides future research directions.

 \vspace{-2mm}
\subsection{Related work}\label{sec:relatedwork}

\paragraph{Prior selection for Bayesian inverse problems.} We focus on prior selection and don't consider the sampling of the posterior, which is a main topic for nonlinear Bayesian inverse problems with a given prior, see, e.g., \cite{dashti2017bayesian,KS05,spantini2015optimal,Stuart10,cui2022unified}.
Prior selection is an important topic in statistical Bayesian modeling, which dates back to \cite{jeffreys1998theory}. This study provides a new class of problems where prior selection is crucial: learning kernels in operators, which is an ill-posed linear inverse problem. Our data-adaptive RKHS prior re-discovers the well-known Zellner's g-prior in \cite{Zellner1980gprior,Agliari1988gprior} when the kernel is finite-dimensional and the basis functions are orthonormal. Importantly, the stable small noise limit in this study provides a new criterion for prior selection, a useful addition to many criteria studied in \cite{Bayarri2012Criteria}. 
 
\paragraph{Regularization in a variational approach.}  The prior is closely related to Tikhonov or ridge regularization in a variation approach. The likelihood function provides a loss function, and the prior often provides a regularization term. Various regularization terms have been studied, including the widely-used Euclidean norm (see, e.g., \cite{gazzola2019ir,hansen1994_regularization_tools,hansen_LcurveIts_a,tihonov1963solution}), the RKHS norm with pre-specified reproducing kernel (see, e.g., \cite{CZ07book,bauer2007regularization}), the total variation norm in \cite{rudin1992nonlinear}, and the data-adaptive RKHS norm in \cite{LAY22,LLA22}. It remains open to compare these norms. Appealingly, the Bayesian approach provides probabilistic tools for analyzing regularizations. Thus, to better understand regularization, it is of interest to study the priors in a Bayesian approach. 


\vspace{-2mm}
\paragraph{Operator learning.} This study focuses on learning the kernels, not the operators. Thus, our focus differs from the focus of the widely-used kernel methods for operator approximation (see, e.g., \cite{owhadi2019kernel,darcy2021learning}) and the operator learning (see, e.g., \cite{HHQ22,HKN22,KLA21,LKN20,LJK19,LJP21}). 
These methods aim to approximate the operator matching the input and output, not to identify the kernel in the operator. 

\vspace{-2mm}
\paragraph{Gaussian process and kernel-based regression.} Selection of the reproducing kernel is an important component in Gaussian process and kernel-based regression (see, e.g.,\cite{belkin2018understand,CZ07book,hofmann2008kernel,Sriperumbudur2011Universality,yuan2010reproducing}). Previous methods often tune the hyper-parameter of a pre-selected class of kernels. Our data-adaptive RKHS prior provides an automatic reproducing kernel, adaptive to data and the model, for these methods. 

\vspace{-3mm}
\paragraph{Learning interacting kernels and nonlocal kernels.} The learning of kernels in operators has been studied in the context of identifying the interaction kernels in interacting particle systems (e.g., \cite{FRT21-GP,he2022numerical,della2022nonparametric,LangLu22,LLMTZ21,LMT21_JMLR,LZTM19pnas,LMT21,mavridis2022learning,messenger2022learning,yao2022mean}) and the nonlocal kernels in homogenization of PDEs (e.g., \cite{LAY22,you2022_DatadrivenPeridynamic,you2021_DatadrivenLearning}). This study is the first to analyze the selection of a prior in a Bayesian approach.

\section{The learning of kernels in operators}
\label{sec:setting}

In this section, we discuss learning kernels in operators as a variational inverse problem that maximizes the likelihood. In this process, we introduce a few key concepts for the Bayesian approach in later sections: the function space for the kernel, the normal operator, the function space of identifiability, and a data-adaptive RKHS.

\subsection{Examples}\label{sec:learningkernels}
We first present a few examples of learning kernels in operators. Note that in these examples, there is little prior information about the kernel. 

 \begin{example}[Kernels in Toeplitz matrices]    \label{exp:Toeplitz_matrix}
Consider the estimation of the kernel $\phi$ in the Toeplitz matrix $R_\phi\in \R^{n\times n}$, i.e., $R_\phi(i,j) = \phi(i-j)$ for all $1\leq i, j\leq n$,
from measurement data $\{(u^k,f^k)\in \R^n\times \R^n\}_{k=1}^N$  by fitting the data to the model
\begin{equation}\label{eq:toeplitz}
R_\phi u +\eta +\xi(u)  = f, \quad \eta \sim \mathcal{N}(0,\sigma_\eta^2 I_n), \quad \spaceX = \spaceY= \R^n, 
\end{equation}
where $\xi(u)$ 
represents an unknown model error. 
We can write the Toeplitz matrix as an integral operator in the form of \eqref{eq:operator_general} with $\Omega = \{1,2,\ldots, n\}$, $g[u](x,y) = u(y)$, and $\mu$ being a uniform discrete measure on $\Omega$. The kernel is a vector $\phi: \calS\to \R^{2n-1}$ with $\calS= \{r_l\}_{l=1}^{2n-1}$ with $r_l= l-n$. 
   \end{example}

\begin{example}[Integral operator]\label{exp:integral} 
   Let $\spaceX= \spaceY= L^2([0,1])$. We aim to find a function $\phi: [-1,1]\to \R$ fitting the dataset in \eqref{eq:data} to the model \eqref{eq:map_R} with an integral operator  
\begin{equation}\label{eq:Int_operator}
R_\phi[u](y) = \int_{0}^1 \phi(y-x) u(x)dx, \quad \forall y\in [0,1]. 
\end{equation}
We assume that $\eta$ is a white noise, that is, $\E[\eta(y)\eta(y')] = \delta(y'-y)$ for any $y,y'\in [0,1]$. 
In the form of the operator in \eqref{eq:operator_general}, we have $\Omega =[0,1]$, $g[u](x,y) = u(x)$, and $\mu$ being the Lebesgue measure. This operator is an infinite-dimensional version of the Toeplitz matrix.  
\end{example}

   \begin{example}[Nonlocal operator]\label{exp:nonlocal} 
 Suppose that we want to estimate a kernel $\phi:\R^d\to\R$ in a model \eqref{eq:map_R} with a nonlocal operator 
\begin{equation*}
R_\phi[u](y) = \int_{\Omega} \phi(y-x) [ u(y) - u(x) ] dx, \quad \forall y\in \Omega \subset \R^d, 
\end{equation*}
 from a given data set as in \eqref{eq:data} with $\spaceX=L^2( \Omega)$ and $\spaceY= L^2( \Omega)$, where $\Omega$ is a bounded set. 
 Such nonlocal operators arise in {\rm\cite{du2012_AnalysisApproximation,you2021_DatadrivenLearning,LAY22}}.   Here $\eta$ is a white noise in the sense that $\E[\eta(y)\eta(y')] = \delta(y-y')$ for any $y,y'\in  \Omega$. This example corresponds to \eqref{eq:operator_general} with $g[u](x,y)= u(y)- u(x)$. Note that even the support of the kernel $\phi$ is unknown. 
\end{example}

\begin{example}[Interaction operator] \label{exp:interaction} 
Let $\spaceX = C^1_0(\R)$ and $\spaceY=L^2(\R)$ and consider the problem of estimating the \emph{interaction kernel} $\phi: \R\to\R$ in the nonlinear operator  
\begin{equation*}
R_\phi[u](y) = \int_{\R} \phi(y-x) [u'(y)u(x) + u'(x)u(y)] dx, \, \quad \forall y\in \R,
\end{equation*}
by fitting the dataset in \eqref{eq:data} to the model \eqref{eq:map_R}. 
This nonlinear operator corresponds to \eqref{eq:operator_general} with $g[u](x,y)= u'(y)u(x)+  u'(x)u(y)$. It comes from the aggression operator $R_\phi[u] = \nabla\cdot  [u \nabla (\Phi*u )]$ in the mean-field equation of interaction particles (see, e.g., {\rm\cite{carrillo2019aggregation,LangLu22}}). 
\end{example}

\subsection{Variational approach}
To identify the kernel, the variational approach seeks a maximal likelihood estimator: 
\begin{equation}\label{eq:lossFn}
\widehat \phi = \argmin_{\phi\in \mH}\calE(\phi), \quad \text{ where } \calE(\phi) = \frac{1}{N\sigma_\eta^2}\sum_{1\leq k\leq N} \|R_\phi[u^k]-f^k\|_{\spaceY}^2,  
\end{equation}
 over a hypothesis space $\mH$, where the loss function $\calE(\phi)$ is the negative log-likelihood of the data \eqref{eq:data} 
under the assumption that the noise $\eta$ in \eqref{eq:map_R} is white. 

The first step is to find a proper function space for $\phi$, in which one can further specify the hypothesis space $\mH$. Clearly, given a data set, we can only identify the kernel where the data provides information. Examples \ref{exp:Toeplitz_matrix}-- \ref{exp:interaction} show that the support of the kernel $\phi$ is yet to be extracted from data.

We set function space for learning $\phi$ to be $L^2(\calS,\rho)$ with $\calS=\{x-y: x,y\in\Omega\}$, where $\rho$ is an empirical probability measure quantifying the exploration of data to the kernel: 
\begin{align}\label{eq:rho_conti}
\rho(dr) =  \frac{1}{ZN}\sum_{1\leq k\leq N}\int_\Omega\int_\Omega \delta(y-x-r)  \left| g[u^k](x,y) \right|\mu(dx)\mu(dy), \quad  r\in \calS. 
 \end{align}
Here $Z$
 is the normalizing constant and  $\delta$ is the Dirac delta function. We call $\rho$ an \emph{exploration measure}. Its support is the region in which the learning process ought to work, and outside of which, we have limited information from the data to learn the function $\phi$. Thus, we can restrict $\calS$ to be the support of $\rho$, and we denote $L^2_\rho:= L^2(\calS,\rho)$ for short. {In other words, the exploration measure is a generalization of the measure $\rho_X$ in nonparametric regression that estimates $Y=f(X)$ from data $\{(x_i,y_i)\}$, where $\rho_X$ is the distribution of $X$ and the data are samples from the joint distribution of $(X,Y)$. }

The next step is to find the minimizer of the loss function. Note that $\calE(\phi)$ is quadratic in $\phi$ since the operator $R_\phi$ depends linearly on $\phi$. We get the minimizer by solving the zero of the (Fr\'echet) derivative of the loss function.  To compute its  derivative, we first introduce a bilinear form $\dbinnerp{\cdot,\cdot}$: $\forall \phi, \psi\in L^2_\rho$,
 \begin{equation}\label{eq:binlinearForm} 
 \begin{aligned} 
\dbinnerp{\phi,\psi}= &\frac{1}{N}\sum_{1\leq k\leq N}\innerp{R_{\phi}[u^k], R_{\psi}[u^k]}_{\spaceY},\\
 = & \frac{1}{N}\sum_{1\leq k\leq N}\int \left[ \int \int \phi(y-x)  \psi(y-z) g[u^k](x,y)g[u^k](z,y) \mu(dx)\mu(dz) \right] \mu(dy)   \\
 = &    \int_\calS \int_\calS \phi(r)\psi(s) \Gbar (r,s) \rho(dr)\rho(ds),  
\end{aligned}
\end{equation}
where the integral kernel $\Gbar$ given by, for $ r,s\in \mathrm{supp}(\rho)$, 
\begin{equation}\label{eq:Gbar}
    \Gbar(r,s) = \frac{G(r,s)}{\rho(r)\rho(s)} \quad \text{ with } G(r,s)= \frac{1}{N}\sum_{1\leq k\leq N}\int  g[u^k](x,r+x)  g[u^k](x,s+x) \mu(dx) ,
\end{equation} 
in which by an abuse of notation,  we also use $\rho(r)$ to denote either the probability of $r$ when $\rho$ defined in \eqref{eq:rho_conti} is discrete or the probability density of $\rho$ when the density exists.

By definition, the bivariate function $\Gbar$ is symmetric and positive semi-definite in the sense that $\sum_{i,j=1}^n c_ic_j G(r_i,r_j)\geq 0$ for any $\{c_i\}_{i=1}^n\subset \R$ and $\{r_i\}_{i=1}^n\subset \calS$. In the following, we assume that the data is continuous and bounded so that $\Gbar$ defines a self-adjoint compact operator which is fundamental for studying identifiability.  This assumption holds under mild regularity conditions on the data $\{u^k\}$ and the operator $R_\phi$. 
\begin{myassumption}[Integrability of $\Gbar$] \label{assumption1}
Assume that $\Omega$ is bounded and $\{g[u^k](x,y)\}$ in \eqref{eq:operator_general} with data $\{u^k\}$ in \eqref{eq:data} are continuous  satisfying $\max_{1\leq k \leq N}\sup_{x,y\in \Omega} |g[u^k](x,y)|<\infty$.  
\end{myassumption}

Under Assumption \ref{assumption1}, the integral operator $\LGbar: L^2_\rho\to L^2_\rho$ 
\begin{equation}\label{eq:LG}
\LGbar \phi (r) =  \int_\calS  \phi(s) \overline{G}(r,s) \rho(s)ds, 
\end{equation}
is a positive semi-definite trace-class operator (see Lemma \ref{lemma:trace-opt}). Hereafter we denote $\{\lambda_i, \psi_i\}$ the eigen-pairs of $\LGbar$ with the eigenvalues in descending order, and assume that the eigenfunctions are orthonormal, hence they provide an orthonormal basis of $L^2_\rho$. 
Furthermore, for any $\phi,\psi\in L^2_\rho$, the bilinear form in \eqref{eq:binlinearForm}  can be written as
\begin{equation}\label{eq:dbinnerp_LG}
\dbinnerp{\phi,\psi} = \innerp{\LGbar \phi,\psi}_{L^2_\rho}= \innerp{\phi,\LGbar \psi}_{L^2_\rho},
\end{equation}
and we can write the loss functional in \eqref{eq:lossFn} as 
\begin{equation}\label{eq:lossFn2}
\begin{aligned}
\calE(\phi) 
 & = \dbinnerp{\phi,\phi} - 2 \frac{1}{N}\sum_{1\leq k\leq N}\innerp{ R_\phi[u^k], f^k}_{\spaceY} + \frac{1}{N}\sum_{1\leq k\leq N}\| f^k \|_\spaceY^2 \\
 & = \innerp{\LGbar \phi,\phi}_{L^2_\rho} -2 \innerp{\phi^\data,\phi}_{L^2_\rho}+C_N^f, 
\end{aligned}
\end{equation}
where  $\phi^\data \in L^2_\rho$ is the Riesz representation of the bounded linear functional: 
\begin{equation}\label{eq:phi_f_N} 
\innerp{\phi^\data,\psi}_{L^2_\rho} = \frac{1}{N}\sum_{1\leq k\leq N}\innerp{ R_\psi[u^k], f^k}_{\spaceY} , \quad \forall \psi\in L^2_\rho.
\end{equation}

Then,  by solving the zero of $ \nabla \calE(\phi) =  2 ( \LGbar \phi - \phi^\data)$, one may obtain the minimizer $\LGbar^{-1}\phi^\data$, provided that it is well-defined. However, it is often ill-defined, e.g., when $\phi^\data$ is not in $\LGbar (L^2_\rho)$ and $\LGbar$ is compact infinite-rank or rank-deficient. Thus, it is important to examine the inversion and the uniqueness of the minimizer, for which we introduce the function space of identifiability in the next section.

\subsection{Function space of identifiability}\label{sec:fsoi}
We define a function space of identifiability (FSOI) when one learns the kernel by minimizing the loss function.  
\begin{definition}\label{def:spaceID}
The function space of identifiability (FSOI) by the loss functional $\calE$ in \eqref{eq:lossFn} is the largest linear subspace of $L^2_\rho$ in which $\calE$ has a unique minimizer. 
\end{definition}
 The next theorem characterizes the FSOI. Its proof is deferred to Appendix \ref{append_ID}.
\begin{theorem}[Function space of identifiability]\label{thm:FSOI}  
Suppose the data in \eqref{eq:data} is generated from the system \eqref{eq:map_R} with a true kernel $\phi_{true}$. Suppose that Assumption {\rm\ref{assumption1}} holds. 
Then, the following statements hold. 
\begin{itemize}\setlength\itemsep{-0mm}
\item[(a)] The function $\phi^\data$ in \eqref{eq:phi_f_N} has the following decomposition: 
\begin{equation}\label{eq:phi_data}
\phi^\data = \LGbar \phi_{true} + \epsilon^{\xi}+ \epsilon^\eta ,
\end{equation}
where $\epsilon^{\xi}$ comes from the model error $\xi$, the random $\epsilon^\eta $ comes from the observation noise and it has a Gaussian distribution $\calN(0, \sigma_\eta^2\LGbar )$, and they satisfy $\forall \psi\in L^2_\rho$
\begin{align*}
&  \innerp{\epsilon^\xi,\psi}_{L^2_\rho}  = \frac{1}{N}\sum_{1\leq k\leq N} \innerp{ R_\psi[u^k], \xi^k }_{\spaceY}, \quad \innerp{\epsilon^\eta, \psi}_{L^2_\rho}= \frac{1}{N}\sum_{1\leq k\leq N}\innerp{ R_\psi[u^k], \eta_k] }_{\spaceY}\,.
\end{align*}
\item[(b)] The Fr\'echet derivative of $\calE(\phi)$ in $L^2_\rho$ is $ \nabla \calE(\phi) =  2 ( \LGbar \phi - \phi^\data)$.
\item[(c)] The function space of identifiability (FSOI) of $\calE$  is $H
= \overline{ \mathrm{span}\{\psi_i\}}_{i: \lambda_i>0}$ with closure in $L^2_\rho$. In particular, if $\phi^\data\in \LGbar(L^2_\rho)$,  the unique minimizer of $\calE(\phi)$ in the FSOI is $\widehat{\phi} = \LGbar^{-1}\phi^\data$, where $\LGbar^{-1}$ is the pseudo-inverse of $\LGbar$.  
Furthermore, if $\phi_{true}\in H$ and there is no observation noise and no model error, we have 
$\widehat{\phi} = \LGbar^{-1}\phi^\data = \phi_{true}$. 
\end{itemize} 
\end{theorem}

Theorem \ref{thm:FSOI} enables us to analyze the ill-posedness of the inverse problems through the operator $\LGbar$ and $\phi^\data$. When $\phi^\data\in \LGbar(L^2_\rho)$, the inverse problem has a unique solution in the FSOI $H$, even when it is underdetermined in $L^2_\rho$ due to $H$ being a proper subspace, which happens when the compact operator $\LGbar$ has a zero eigenvalue. 
However, when $\phi^\data\notin \LGbar(L^2_\rho)$, the inverse problem $\nabla \calE =0$ has no solution in $L^2_\rho$ because  $\LGbar^{-1}\phi^\data$ is undefined. According to \eqref{eq:phi_data}, this happens in one or more of the following scenarios:  
\begin{itemize}
\item  when the model error leads to $\epsilon^\xi\notin  \LGbar(L^2_\rho)$. 
\item when the observation noise leads to $\epsilon^\eta\notin \LGbar(L^2_\rho)$. In particular, since $\epsilon^\eta$ is Gaussian $\calN(0,\LGbar)$, it has the Karhunen--Lo\`eve expansion $\epsilon^\eta=\sum_i \lambda_i^{1/2} \epsilon^\eta_i \psi_i$ with $ \epsilon^\eta_i$ being i.i.d.~$\calN(0,1)$. Then, $\LGbar^{-1}\epsilon^\eta = \sum_i \lambda_i^{-1/2} \epsilon^\eta_i \psi_i$, which diverges a.s.~if $\sum_{i:\lambda_i>0} \lambda_i^{-1}$ diverges.  Thus, we have $\epsilon^\eta\notin  \LGbar(L^2_\rho) $ a.s.~when $\sum_{i:\lambda_i>0} \lambda_i^{-1}$ diverges. 
 \end{itemize}
Additionally, $\phi^\data$ only encodes information of $\phi_{true}^H$, 
and it provides no information about $ \phi_{true}^\perp$, where $\phi_{true}^H$ and $ \phi_{true}^\perp$ form an orthogonal decomposition $\phi_{true} = \phi_{true}^H + \phi_{true}^\perp \in H \oplus H^\perp$. In other words, the data provides no information to recover $ \phi_{true}^\perp$. 

 As a result, it is important to avoid absorbing the errors outside of the FSOI when using a Bayesian approach or a regularization method to mitigate the ill-posedness.

\subsection{A data-adaptive RKHS}\label{sec:dartr}  
The RKHS with $\Gbar$ as a reproducing kernel is a dense subspace of the FSOI. Hence, when using it as a prior, which we will detail in later sections, one can filter out the error outside the FSOI and ensure that the learning takes place inside the FSOI. 

The next lemma is an operator characterization of this RKHS (see, e.g., \cite[Section 4.4]{CZ07book}). Its proof can be found in \cite[Theorem 3.3]{LLA22}.  

\begin{lemma}[A data-adaptive RKHS] \label{lemma:sidaRKHS}
Suppose that Assumption {\rm\ref{assumption1}} holds. 
 Then,  the following statements hold. 
\begin{itemize}\setlength\itemsep{-0mm}
\item[(a)] The RKHS $H_G$ with $\Gbar$ in \eqref{eq:Gbar} as reproducing kernel satisfies $H_G = \LGbar^{1/2} (L^2_\rho )$ and its inner product satisfies  
\begin{equation}\label{eq:HG}
\innerp{\phi,\psi}_{H_G} = \innerp{\LGbar^{-1/2}\phi,\LGbar^{-1/2}\psi}_{L^2_\rho}, \quad \forall \phi,\psi\in H_G.
\end{equation}
\item[(b)] Denote the eigen-pairs of $\LGbar$ by $\{\lambda_i,\psi_i\}_{i}$ with $\{\psi_i\}$ being orthonormal.  
Then, for any $\phi= \sum_i c_i \psi_i\in L^2_\rho$, we have 
\begin{equation}\label{eq:norms}
\dbinnerp{\phi,\phi} = \sum_i \lambda_i c_i^2, \quad \|\phi\|^2_{L^2_\rho} = \sum_i  c_i^2, \quad  \|\phi\|^2_{H_G} = \sum_{i:\lambda_i>0} \lambda_i^{-1} c_i^2, 
 \end{equation}
where the last equation is restricted to $\phi\in H_G$. 
\end{itemize}
\end{lemma}

 The above RKHS has been used for regularization in \cite{LAY22,LLA22}. The regularizer, named DARTR, regularizes the loss by the norm of this RKHS,
 \begin{equation}\label{eq:dartrLoss}
\calE_\lambda(\phi) = \calE(\phi)+\lambda \|\phi\|_{H_G}^2 = \innerp{(\LGbar+\lambda \LGbar^{-1}) \phi,\phi}_{L^2_\rho} -2 \innerp{\phi^\data,\phi}_{L^2_\rho}+C_N^f.
\end{equation}
Selecting the optimal hyper-parameter $\lambda_*$ by the L-curve method, it leads to the estimator 
\begin{equation}\label{eq:dartrEst}
\widehat \phi_{H_G} = (\LGbar^2+ \lambda_{*} I_H)^{-1} \LGbar \phi^\data, 
\end{equation}
where $I_H$ is the identity operator on $H$. Since the RKHS is a subset of the FSOI with more regular elements, DARTR ensures that the estimator is in the FSOI and is regularized.

We summarize the key terminologies and notations in this section in Table \ref{tab:term}.
\begin{table}[ht]
\caption{Notations for learning kernels in operators in Sect.\ref{sec:setting}.} \vspace{-3mm}
\label{tab:term}
 \begin{center}
 \begin{tabularx}{\textwidth}
 { X|X 
    }
 \toprule
 \hline
 \textbf{Notation} & \textbf{Meaning} \\
 \hline
  Exploration measure $\rho$  &   
    A probability measure quantifying the data's exploration to the kernel \eqref{eq:rho_conti}. \\  \hline
  $L^2(\calS,\rho)$ or $L^2_\rho$       & Function space of learning \\ \hline
 $\LGbar$, $\{(\lambda_i,\psi_i)\}_i$,   $\LGbar^{-1}$    & The normal operator in \eqref{eq:LG}, its spectral-decomposition and pseudo-inverse \\ \hline
 $ \calE(\phi) = \innerp{\LGbar \phi,\phi}_{L^2_\rho} -2 \innerp{\phi^\data,\phi}_{L^2_\rho}+C_N^f$ & Loss function from the likelihood in \eqref{eq:lossFn2}  \\ \hline
 FSOI  $H= \mathrm{span}\{\psi_i\}_{\lambda_i>0}$   & Function space of identifiability, in which $\calE$ has a unique minimizer.\\ \hline
\bottomrule
 \end{tabularx} 
 \end{center}\vspace{-3mm}
\end{table}

\section{Bayesian inversion and the risk in a non-degenerate prior}\label{sec:Bayes}
The Bayesian approach overcomes the ill-posedness by introducing a prior, so the posterior is stable under perturbations. Since little prior information is available about the kernel, it is common to use a non-degenerate prior to ensure the well-posedness of the posterior. However, we will show that a fixed non-degenerate prior may lead to a posterior with a divergent mean in the small noise limit. These discussions promote the data-adaptive prior in the next section.

\subsection{The Bayesian approach}\label{sec:Bayes_def}

In this study, we focus on Gaussian priors, so the posterior is also a Gaussian measure in combination with a Gaussian likelihood. Also, without loss of generality, we assume that the prior is centered.  
Recall that the function space of learning is $L^2_\rho$ defined in \eqref{eq:rho_conti}. For illustration, we first specify the prior and posterior when the space $L^2_\rho$ is finite-dimensional, then discuss them in the infinite-dimensional case. We follow the notations in \cite{Stuart10} and \cite{dashti2017bayesian}. 

\paragraph{Finite-dimensional case.}
Consider first that the space $L^2_\rho$ is finite-dimensional, i.e., $\calS = \{r_1,\ldots, r_{d}\}$, as in Example \ref{exp:Toeplitz_matrix}. Then, the space $L^2_\rho$ is equivalent to $\R^{d}$ with norm satisfying $\|\phi\|^2= \sum_{i=1}^{d} \phi(r_i)^2 \rho(r_i)$.   Also, assume that space $\spaceY$ is finite-dimensional, and the measurement noise in \eqref{eq:data} is Gaussian $\mathcal{N}(0,\sigma_\eta^2 I)$. Since $\phi$ is finite-dimensional, we write the prior, the likelihood, and the posterior in terms of their probability densities with respect to the Lebesgue measure. 
\begin{itemize}[leftmargin=*]
\item {\bf Prior} distribution, denoted by $\calN(0,\calQ_0)$, with density  
\[ \frac{d\prior(\phi)}{d\phi} \propto e^{-\frac{1}{2}\innerp{\phi,\calQ_0^{-1}\phi}_{L^2_\rho}},\]
 where $\calQ_0$ is a strictly positive matrix, so that the prior is a non-degenerate measure. 
\item {\bf Likelihood} distribution of the data with density 
\begin{equation}\label{eq:likelihood}
\frac{d\likelihood(\phi)}{d\phi} \propto \exp\left(-\frac{1}{2\sigma_{\eta}^{2}} \calE(\phi)\right) =\exp\left(-\frac{1}{2\sigma_{\eta}^{2}}  \left[ \innerp{\LGbar \phi,\phi}_{L^2_\rho} -2 \innerp{\phi^\data,\phi}_{L^2_\rho}+C_N^f \right] \right),
\end{equation}
where $\calE(\phi)$ is the loss function defined in \eqref{eq:lossFn} and the equality follows from \eqref{eq:lossFn2}. 
 Note that this distribution is a non-degenerate Gaussian $\calN(\LGbar^{-1}\phi^\data, \sigma_\eta^{2}  \LGbar^{-1} )$ when $\LGbar^{-1} $ exists, and it can be ill-defined when $\LGbar$ has a zero eigenvalue.
\item {\bf Posterior} distribution with density proportionating to the product of the prior and likelihood densities, 
\begin{align}
\frac{d\posterior(\phi)}{d\phi}&\propto \exp\left(-\frac{1}{2}[ \sigma_{\eta}^{-2} \calE(\phi) + \innerp{\calQ_0^{-1}\phi,\phi}_{L^2_\rho} ] \right) \nonumber \\
	&= \exp\left(-\frac{1}{2} \Big[\sigma_{\eta}^{-2} \big(\innerp{\LGbar \phi,\phi}_{L^2_\rho} -2 \innerp{\phi^\data,\phi}_{L^2_\rho}+C_N^f \big) + \innerp{\phi,\calQ_0^{-1}\phi}_{L^2_\rho}\Big] \right)\nonumber \\
	&= \exp\left(-\frac12 \innerp{\calQ_1^{-1}(\phi-\mu_1),\phi-\mu_1}+C_N^f \right)\label{eq:posterior_density}
\end{align}
with the constant term $C_N^f$ may change from line to line and 
\begin{equation}\label{eq:posterior}
 \mu_1 = (  \LGbar+ \sigma_{\eta}^{2} \calQ_0^{-1} )^{-1} \phi^\data =\sigma_{\eta}^{-2}\calQ_1 \phi^\data, \, \text{and } \calQ_1= \sigma_{\eta}^{2} (  \LGbar + \sigma_{\eta}^{2} \calQ_0^{-1} )^{-1} .
\end{equation}  
Thus, $\posterior(\phi)$ is a Gaussian measure $\calN(\mu_1,\calQ_1)$. 
\end{itemize}
 
The Bayesian approach is closely related to Tikhonov regularization (see, e.g., \cite{Lehtinen}). Note that a Gaussian prior corresponds to a regularization term $\mathcal{R}(\phi) = \innerp{\phi,\calQ_0^{-1}\phi}_{L^2_\rho}$, the negative log-likelihood is the loss function, and the posterior corresponds to the penalized loss: 
\[
- 2\sigma_\eta^{2} \log \posterior(\phi)  =  \calE(\phi) +\lambda \innerp{\phi,\calQ_0^{-1}\phi}_{L^2_\rho}  \, \text{ with }  \lambda= \sigma_\eta^{2}. 
\]
In particular, the \emph{maximal a posteriori}, MAP in short, which agrees with the posterior mean $\mu_1$, is the minimizer of the penalized loss using a penalty term $\sigma_\eta^{2} \innerp{\phi,\calQ_0^{-1}\phi}_{L^2_\rho} $. The difference is that a regularization approach selects the hyper-parameter according to data. 

\paragraph{Infinite-dimensional case.} When space $L^2_\rho$ is infinite-dimensional, i.e., the set $\calS$ has infinite elements, we use the generic notion of Gaussian measures on Hilbert spaces, see Appendix \ref{sec:gaussian_Hilbert} for a brief review. Since there is no longer a Lebesgue measure on the infinite-dimensional space,  the prior and the posterior are characterized by their means and covariance operators. We write the prior and the posterior as follows: 

\begin{itemize}[leftmargin=*]
\item {\bf Prior} $\calN(0,\calQ_0)$, where $\calQ_0$ is a strictly positive trace-class operator on $L^2_\rho$; 
\item {\bf Posterior} $\calN(\mu_1,\calQ_1)$, whose Radon--Nikodym derivative with respect to the prior is   
\begin{equation}\label{eq:likelihood_hilbert}
\frac{d\posterior}{d\prior} \propto  \exp(-\frac{1}{2}\sigma_{\eta}^{-2} \calE(\phi)) =\exp\left(-\frac{1}{2}\sigma_\eta^{-2}  \left[ \innerp{\LGbar \phi,\phi}_{L^2_\rho} -2 \innerp{\phi^\data,\phi}_{L^2_\rho}+C_N^f \right] \right),
\end{equation}
which is the same as the likelihood in \eqref{eq:likelihood}. Its mean and covariance are given as in \eqref{eq:posterior}. 
\end{itemize}

Note that unlike the finite-dimensional case, it is problematic to write the likelihood distribution as $\calN(\LGbar^{-1}\phi^\data, \sigma_\eta^{-2}  \LGbar^{-1} )$, because the operator $\LGbar^{-1}$ is unbounded and $\LGbar^{-1}\phi^\data$ may be ill-defined.

\subsection{The risk in a non-degenerate prior}\label{sec:fixed_prior}

The prior distribution plays a crucial role in Bayesian inverse problems. To make the ill-posed inverse problem well-defined, it is often set to be a non-degenerate measure (i.e., its covariance operator $\calQ_0$ has no zero eigenvalues). It is fixed in many cases and not adaptive to data. Such a non-degenerate prior works well for an inverse problem whose function space of identifiability does not change with data. However, in the learning of kernels in operators, a non-degenerate prior has a risk of leading to a catastrophic error: the posterior mean may diverge in the small observation noise limit, as we show in the Proposition \ref{thm:risk_prior}.    
  \begin{myassumption}\label{assumption2}
Assume that the operator $\LGbar$ is finite rank and commutes with the prior covariance $\calQ_0$ and assume the existence of error outside of the FSOI as follows. 
  \begin{itemize}
\item[$\assump$] 
The operator $\LGbar$ in \eqref{eq:LG} has zero eigenvalues. Let $\lambda_{K+1}=0$ be the first zero eigenvalue, where $K$ is less than the dimension of $L^2_\rho$. As a result, the FSOI is $H=\mathrm{span}\{\psi_i\}_{i=1}^{K}$. 
\item[$\assumpp$]
The covariance of the prior $\calN(0,\calQ_0)$ satisfies $\calQ_0\psi_i = r_i \psi_i$ with $r_i>0$ for all $i$, where $\{\psi_i\}_{i}$ are orthonormal eigenfunctions of $\LGbar$. 
\item[$\assumppp$] 
The term $\epsilon^\xi $ in \eqref{eq:phi_data}, which represents the model error, is outside of the FSOI, i.e., $\epsilon^\xi    = \sum_i \epsilon^\xi_i   \psi_i $ has a component $\epsilon^\xi_{i_0} \neq 0$ for some $i_0 >K$. 
\end{itemize}
 \end{myassumption}
  
 Assumptions (A1-A2) are common in practice. Assumption (A1) holds because the operator $\LGbar$ is finite rank when the data is discrete, and it is not full rank for under-determined problems. It is natural to assume the prior has a full rank covariance $\calQ_0$ as in (A2). We assume that $\calQ_0$ commutes with $\LGbar$ for simplicity and one can extend it to the general case as in the proof of \cite[{Theorem 2.25, Feldman–Hajek theorem}]{da2014stochastic}.  Assumption (A3), which requires $\phi^\data$ to be outside the range of $\LGbar$, holds when the regression vector $\bbar$ is outside the range of the regression matrix $\Abar$ in \eqref{eq:Abarbbar}, see Sect.\ref{sec:Toeplitz}--\ref{sec:integral-operator} for more discussions. 
  
\begin{proposition}[Risk in a fixed non-degenerate prior]\label{thm:risk_prior}  
A non-degenerate prior has the risk of leading to a divergent posterior mean in the small noise limit. Specifically, under Assumption {\rm\ref{assumption2}}, the posterior mean $\mu_1$ in \eqref{eq:posterior} diverges as $\sigma_\eta\to 0$. 
\end{proposition}
\begin{proof}[Proof of Proposition \ref{thm:risk_prior}.]
Recall that conditional on the data, the observation noise-induced term $\epsilon^\eta$ in \eqref{eq:phi_data} has a distribution $\calN(0,\sigma^2_\eta \LGbar)$. Thus, in the orthonormal basis $\{\psi_i\}$, we can write $\epsilon^{\eta} = \sigma_\eta \sum_{i:\lambda_i>0} \lambda_i^{1/2} \epsilon^{\eta}_i\psi_i$, where $\{\epsilon^{\eta}_i\}$ are i.i.d.~$\calN(0,1)$ random variables. Additionally, write the true kernel as $\phi_{true} =  \sum_{i} \phi_{true,i}\psi_i$, where $\phi_{true,i} = \innerp{\phi_{true},\psi_i}_{L^2_\rho}$ for all $i$. Note that $\phi_{true}$ does not have to be in the FSOI. Combining these facts, we have 
\begin{equation}\label{eq:phi_d_eig}
\phi^\data = \sum_{i=1}^{\infty} \phi^\data_i \psi_i, \, \text{ with } \phi^\data_i  = \lambda_i \phi_{true,i} +\sigma_\eta  \lambda_i^{1/2} \epsilon^{\eta}_i + \epsilon^\xi_i  . 
\end{equation}
Then, the posterior mean $\mu_1 = (  \LGbar+ \sigma_{\eta}^{2} \calQ_0^{-1} )^{-1} \phi^\data $  in \eqref{eq:posterior}  becomes 
\begin{align} 
\mu_1  
&= \sum_{i=1}^{\infty}  \left(\lambda_i+\sigma_{\eta}^{2}  r_i^{-1}  \right)^{-1} \phi^\data_i \psi_i  
=  \sum_{i=1}^K \left(\lambda_i+\sigma_{\eta}^{2}  r_i^{-1}  \right)^{-1}
 \phi^\data_i \psi_i  
+  \sum_{i> K} \sigma_{\eta}^{-2}  r_i \epsilon^\xi_i  \psi_i. \label{eq:m1}
\end{align}
Thus, the posterior mean $\mu_1$ is contaminated by the model error outside the FSOI, i.e., the part with components $\epsilon^\xi_i $ with $i>K$. It diverges when $\sigma_\eta\to 0$ because 
\[
\lim_{\sigma_\eta\to 0} \left( \mu_1- \sum_{i> K} \sigma_{\eta}^{-2}  r_i \epsilon^\xi_i  \psi_i\right) = \sum_{1\leq i\leq K} \left( \phi_{true,i} + \lambda_i^{-1} \epsilon^\xi_i  \right)\psi_i, 
\]
and $ \sum_{i> K} \sigma_{\eta}^{-2}  r_i \epsilon^\xi_i  \psi_i$ diverges.  
\end{proof} 

On the other hand, one may adjust the prior covariance by the standard deviation of the noise in the hope of removing the risk of divergence. However, the next proposition shows that such a noise-adaptive non-degenerate prior will have a biased small noise limit. 

\begin{proposition}[Risk in a noise-adaptive non-degenerate prior]\label{thm:risk_prior2} 
Let the prior be $\calN(0,\lambda \calQ_0)$, where $\lambda=C_0\sigma_\eta^{2\beta}$ with $C_0\neq 0$ and $\beta\geq 0$.  Suppose that Assumption {\rm\ref{assumption2}} holds. Then,  the corresponding posterior mean $\mu_1^{\sigma_\eta}$ either blows up or is biased, satisfying 
	\begin{align*}
		\lim_{\sigma_\eta\to 0}\mu_1^{\sigma_\eta} =
		\begin{cases}
			\infty\,, & \beta<1\,; \\
			(  \LGbar+ C_0^{-1} \calQ_0^{-1} )^{-1} \phi^\data\,, & \beta=1\,; \\
			0\,, & \beta>1\,.
		\end{cases}
	\end{align*}
\end{proposition}
\begin{proof}
	Under the prior $\calN(0,\lambda \calQ_0)$ with $\lambda=C_0\sigma_\eta^{2\beta}$, we have the posterior mean
	\begin{align*} 
	\mu_1^{\sigma_\eta}  &= (  \LGbar+ \sigma_{\eta}^{2} \lambda^{-1} \calQ_0^{-1} )^{-1} \phi^\data = \sum_{i=1}^{\infty}  \left(\lambda_i+\sigma_{\eta}^{2}  \lambda^{-1} r_i^{-1}  \right)^{-1} \phi^\data_i \psi_i  \\
	&=  \sum_{i=1}^K \left(\lambda_i+\sigma_{\eta}^{2} \lambda^{-1}  r_i^{-1}  \right)^{-1}  \phi^\data_i \psi_i  
	+  \sum_{i> K} \sigma_{\eta}^{-2} \lambda r_i \epsilon^\xi_i  \psi_i \\
	&= \sum_{i=1}^K \left(\lambda_i+C_0^{-1} \sigma_{\eta}^{2-2\beta}   r_i^{-1}  \right)^{-1} \phi^\data_i \psi_i + \sum_{i> K} C_0\sigma_{\eta}^{2\beta-2}  r_i \epsilon^\xi_i  \psi_i\,.
	\end{align*}
	Then, the limits for $\beta>1$, $\beta=1$ and $\beta<1$ follow directly. Note that when $\beta\geq 1$, the limits are biased, not recovering the identifiable component  $ \LGbar^{-1}\phi^\data= \sum_{i=1}^K \lambda_i^{-1} \phi^\data_i \psi_i$ even when the model error $\sum_{i=1}^K\epsilon^\xi_i\psi_i$ vanishes. 
\end{proof}

Propositions \ref{thm:risk_prior} and \ref{thm:risk_prior2} highlight that the risk of a non-degenerate prior comes from the error outside the data-adaptive FSOI. Importantly, by definition of the FSOI, there is no signal to be recovered outside the FSOI. Thus, it is important to design a data-adaptive prior to restrict the learning to take place inside the FSOI. 

\section{Data-adaptive RKHS prior}\label{sec:DAprior}
 
We propose a data-adaptive prior to filter out the error outside of the FSOI, so that its posterior mean always has a small noise limit. In particular, the small noise limit converges to the identifiable part of the true kernel when the model error vanishes. Additionally, we show that this prior, even with a sub-optimal $\lambda_*$,  outperforms a large class of fixed non-degenerate priors in the quality of the posterior. 

\subsection{Data-adaptive prior and its posterior}
We first introduce the data-adaptive prior and specify its posterior, which will remove the risk in a non-degenerate prior as shown in Propositions \ref{thm:risk_prior}--\ref{thm:risk_prior2}. This prior is a Gaussian measure with a covariance from the likelihood.

Following the notations in Sect.\ref{sec:learningkernels},  the operator $\LGbar$ is a data-dependent positive definite trace-class operator on $L^2_\rho$, and we denote its eigen-pairs by $\{\lambda_i,\psi_i\}_{i\geq 1}$ with the eigenfunction forming an orthonormal basis of $L^2_\rho$. Then, as characterized in Theorem \ref{thm:FSOI} and Lemma \ref{lemma:sidaRKHS}, the data-dependent FSOI and RKHS are 
\begin{equation}\label{eq:H-HG}
H=\overline{\mathrm{span}\{\psi_i\}_{\lambda_i>0}}^{\|\cdot\|_{L^2_\rho}},\quad 
H_G =\overline{\mathrm{span}\{\psi_i\}_{\lambda_i>0}}^{\|\cdot\|_{H_G}},  
\end{equation}
where the closure of $H_G$ is with respect to the norm $\|\phi\|_{H_G}^2= \sum_{i:\lambda_i>0}\lambda_i^{-1} \innerp{\phi,\psi_i}_{L^2_\rho}^2$. 
Note that those two spaces are the same vector space but with different norms. {These two spaces are different unless the operator $\LGbar$ is finite rank (e.g., the cases in Section \ref{sec:fixed_prior}).} They are proper subspaces of $L^2_\rho$ when the operator $\LGbar$ has a zero eigenvalue.

\begin{definition}[Data-adaptive RKHS prior] Let  $\LGbar$ be the operator defined in \eqref{eq:LG}.  The data-adaptive prior is a Gaussian measure with mean and covariance defined by 
\begin{equation} \label{eq:new_prior}
\prior^\data = \calN(\mu_0^\data, \calQ_0^\data):\quad \mu_0^\data = 0;  \,\, \calQ_0^\data  =\lambda_*^{-1} \LGbar, 
\end{equation}
where the hyper-parameter $\lambda_*> 0$ is determined adaptive to data. 
\end{definition}
In practice, we select the hyper-parameter $\lambda_*\geq 0$ adaptive to data by the L-curve method in \cite{hansen_LcurveIts_a}, which is effective in reaching an optimal trade-off between the likelihood and the prior (see Sect.\ref{sec:DAprior-comput} for more details). We call this prior an RKHS prior because its covariance operator $\LGbar$'s \emph{Cameron-Martin space} is the RKHS  $H_G$ (see, e.g., \cite[Section 1.7]{da2006introduction} and a brief review of the Gaussian measures in Sect.\ref{sec:gaussian_Hilbert}).

This data-adaptive prior is a Gaussian distribution with support in the FSOI $H$ in \eqref{eq:H-HG}. When $H$ is finite-dimensional, its probability density in $H$ is
\[ \frac{d\prior^\data(\phi)}{d\phi} \propto e^{-\frac{1}{2}\innerp{\phi-\mu_0^\data,[\calQ_0^\data]^{-1}  (\phi-\mu_0^\data) }_{L^2_\rho}} = e^{-\frac{1}{2}\innerp{\phi,\lambda_* \LGbar^{-1}  \phi }_{L^2_\rho}}, \quad \forall \phi\in H\]
by definitions \eqref{eq:new_prior}, where $\LGbar^{-1}$ is pseudo-inverse. Combining with the likelihood \eqref{eq:likelihood}, the posterior becomes 
\begin{subnumcases}{\label{eq:posterior_D}}
	\calQ_1^\data =\sigma_\eta^2 ( \LGbar + \sigma_\eta^2 \lambda_* \LGbar^{-1}   )^{-1} = \sigma_\eta^2 ( \LGbar^2 + \sigma_\eta^2  \lambda_* I_H  )^{-1} \LGbar \,; \label{eq:posterior_D_Q} \\
	\mu_1^\data = \sigma_\eta^{-2}\calQ_1^\data \phi^\data=( \LGbar^2 + \sigma_\eta^2 \lambda_* I_H )^{-1} \LGbar \phi^\data.\label{eq:posterior_D_mu}
\end{subnumcases}
We emphasize that the operator $\LGbar^{-1}$ restricts the posterior to be supported on the FSOI $H$, where the inverse problem is well-defined. The posterior mean in \eqref{eq:posterior_D_mu} coincides with the minimum norm least squares estimator if there is no regularization, i.e., $\lambda_*=0$.

When $H$ is infinite-dimensional, the above mean and covariance remain valid, following similar arguments based on the likelihood ratio in \eqref{eq:likelihood_hilbert}.

In either case, the posterior is a Gaussian distribution whose support is $H$, and it is degenerate in $L^2_\rho$ if $H$ is a proper subspace of $L^2_\rho$. 
In short, the data-adaptive prior is a Gaussian distribution supported on the FSOI with a hyper-parameter adaptive to data.  Both the prior and posterior are degenerate when the FSOI is a proper subspace of $L^2_\rho$. 

Table \ref{tab:priors_posteriors} compares the posteriors of the non-degenerate prior and the data-adaptive prior.

\begin{table}[htp] 
	\begin{center} 
		\caption{  \text{Priors and posteriors on $L^2_\rho$.  }} \label{tab:priors_posteriors}
		\begin{tabular}{l l   l l }		\toprule 
                 Gaussian measure                           & Mean  & Covariance  \\  \hline
                 \multicolumn{2}{l}{\emph{Fixed non-degenerate prior and its posterior}} & & \\
	         $\prior  = \calN(\mu_0,\calQ_0)$   &   $\mu_0=0$  & $\calQ_0 $\\ 
	         $\posterior = \calN(\mu_1,\calQ_1)$      &   $\mu_1 = \sigma_\eta^{-2} \calQ_1  \phi^\data $  & $\calQ_1 = \sigma_{\eta}^{2} (  \LGbar + \sigma_{\eta}^{2} \calQ_0^{-1} )^{-1}  $ \vspace{2mm} \\
	           \hline  
	           \multicolumn{2}{l}{\emph{Data-adaptive prior and its posterior} } & & \\
	         $\prior^\data =\calN(\mu_0^\data,\calQ_0^\data)$   &   $\mu_0^\data=0$  & $\calQ_0^\data = \lambda_*^{-1} \LGbar $\\ 
	        $\posterior^\data = \calN(\mu_1^\data,\calQ_1^\data)$      &   $\mu_1^\data=  \sigma_\eta^{-2} \calQ_1^\data  \phi^\data$  & $\calQ_1^\data =\sigma_\eta^2 ( \LGbar^2 + \sigma_\eta^2 \lambda_*I_H   )^{-1} \LGbar $ \rule{0pt}{2.5ex} \\
	 			\bottomrule	  
		\end{tabular}  \vspace{-4mm}
	\end{center}
\end{table}

\subsection{Quality of the posterior and its MAP estimator} \label{sec:quality_posterior}
The data-adaptive prior aims to improve the quality of the posterior. Compared with a fixed non-degenerate prior, we show that the data-adaptive prior improves the quality of the posterior in three aspects: (1) it improves the stability of the MAP estimator so that the MAP estimator always has a small noise limit;  (2)  it improves the accuracy of the MAP estimator by reducing the expected mean square error; and (3) it reduces the uncertainty in the posterior in terms of the trace of the posterior covariance. 

We show first that the posterior mean always has a small noise limit, and the limit converges to the projection of the true function in the FSOI when the model error vanishes. 
\begin{theorem}[Small noise limit of the MAP estimator] \label{thm:MAP_stability}
Suppose that Assumption {\rm \ref{assumption2} (A1-A2)} holds. Then, the posterior mean in \eqref{eq:posterior_D} with the data-adaptive prior \eqref{eq:new_prior}  always has a small noise limit. In particular, its small noise limit converges to the projection of true kernel in the FSOI $H$ in \eqref{eq:H-HG} when the model error $\sum_{i=1}^K\epsilon^\xi_i\psi_i$ vanishes. 
\end{theorem}

 \begin{proof}
 The claims follow directly from the definition of the new posterior mean in \eqref{eq:posterior_D} and the decomposition in Eq.~\eqref{eq:phi_d_eig}, which says that $\phi^\data = \sum_{i} \phi^\data_i \psi_i$ with  $\phi^\data_i  = \lambda_i \phi_{true,i} +\sigma_\eta  \lambda_i^{1/2} \epsilon^{\eta}_i + \epsilon^\xi_i $. Recall that $\mu_1^\data = ( \LGbar^2 + \sigma_\eta^2 \lambda_* I_H )^{-1}\LGbar \phi^\data$, and if $ i\geq K+1$, we have    $ ( \LGbar^2 + \sigma_\eta^2 \lambda_* I_H )^{-1}\LGbar  \psi_{i}=\lambda_i(\lambda_i^2+\sigma_\eta^2 \lambda_*)^{-1} \psi_{i} =0$. 
 Thus, we can write 
 \begin{align}\label{eq:m1_data}
 \mu_1^\data  &= ( \LGbar^2 + \sigma_\eta^2 \lambda_* I_H )^{-1}\LGbar \phi^\data =\sum_{i=1}^K \lambda_i(\lambda_i^2+\sigma_\eta^2 \lambda_*)^{-1} \psi_{i} \phi^\data_i ,
 \end{align}
 since $\{\psi_i\}_{i}$ are orthonormal eigenfunctions of $\LGbar$ with eigenvalues $\{\lambda_i\}_{i}$.   
 Thus,  the small noise limit exists and is equal to
  \begin{align*}
 	\lim_{\sigma_\eta\to 0}\mu_1^\data &=\lim_{\sigma_\eta\to 0}\sum_{1\leq i\leq K}\lambda_i\left(\lambda_i^2+\sigma_{\eta}^{2} \lambda_*  \right)^{-1} \phi^\data_i \psi_i \\ 
 	&= \lim_{\sigma_\eta\to 0}\sum_{1\leq i\leq K}\lambda_i \left(\lambda_i^2+\sigma_{\eta}^{2} \lambda_*  \right)^{-1}(\lambda_i \phi_{true,i} +\sigma_\eta  \lambda_i^{1/2} \epsilon^{\eta}_i + \epsilon^\xi_i) \psi_i\\
 	&=\sum_{i=1}^K \lambda_i^{-1} \big( \lambda_i \phi_{true,i}  + \epsilon^\xi_i \big) \psi_i
 	=  \sum_{i=1}^K \left( \phi_{true,i} + \lambda_i^{-1} \epsilon^\xi_i \right) \psi_i \,.
 \end{align*}  
 Furthermore, as the model error $\|\sum_{i=1}^K\epsilon^\xi_i\psi_i\|^2_{L^2_\rho}= \sum_{i=1}^K |\epsilon^\xi_i  |^2\to 0$, this small noise limit converges to $\sum_{i=1}^K \phi_{true,i} \psi_i$,  the projection of $\phi_{true}$ in the FSOI. 
 \end{proof}

We show next that the data-adaptive prior leads to a MAP estimator more accurate than the non-degenerate prior's. 
\begin{theorem}[Expected MSE of the MAP estimator] \label{thm:MSE}
Suppose that Assumption {\rm \ref{assumption2} (A1-A2)} holds.  Assume in addition that $\max_{i\leq K}\{\lambda_i r_i^{-1}\}\leq \lambda_*\leq 1$. Then, the expected mean square error of the MAP estimator of the data-adaptive prior  is smaller than the non-degenerate prior's, i.e., 
   \begin{equation}\label{eq:MSEineq}
 \E_{\pi_0^\data} \E_\eta \Big[\|\mu_1^\data-\phi_{true}\|_{L^2_\rho}^2\Big] \leq \E_{\pi_0}\E_\eta\Big[\|\mu_1-\phi_{true}\|_{L^2_\rho}^2\Big], 
\end{equation}
where the equality holds only when the two priors are the same. Here $\E_{\pi_0^\data}$ and $\E_{\pi_0}$ denote expectation with $\phi_{true}\sim \pi_0^\data$ and $\phi_{true}\sim \pi_0$, respectively. 
\end{theorem}

\begin{proof}
Note that from \eqref{eq:m1} and \eqref{eq:m1_data}, we have 
\begin{align*}
\mu_1^\data -\phi_{true} 
&=\sum_{1\leq i\leq K} \psi_i\left(\lambda_i+\sigma_{\eta}^{2} \lambda_*\lambda_i^{-1}  \right)^{-1} [\sigma_\eta  \lambda_i^{1/2} \epsilon^{\eta}_i-\sigma_{\eta}^{2} \lambda_* \lambda_i^{-1} \phi_{true,i} + \epsilon^\xi_i  ] + \sum_{i>K}\phi_{true,i}, \\
\mu_1 -\phi_{true} 
&=\sum_{i\geq 1} \psi_i\left(\lambda_i+\sigma_{\eta}^{2} r_i^{-1}  \right)^{-1} [\sigma_\eta  \lambda_i^{1/2} \epsilon^{\eta}_i- \sigma_{\eta}^{2} r_i^{-1} \phi_{true,i} + \epsilon^\xi_i  ]. 
\end{align*}
Recall that $\{\epsilon^{\eta}_i\}$ and $\{ \phi_{true,i}\}$ are independent centered Gaussian with $\epsilon^{\eta}_i\sim\calN(0,1)$, $\phi_{true,i}\sim \calN(0,\lambda_i)$ when $\phi_{true}\sim \prior^\data$, and $\phi_{true,i}\sim \calN(0,r_i)$ when $\phi_{true}\sim \prior$. 
 Then, the expectations of the MSEs $\E_\eta \Big[\|\mu_1^\data-\phi_{true}\|_{L^2_\rho}^2\Big]$ and $\E_\eta \Big[\|\mu_1-\phi_{true}\|_{L^2_\rho}^2\Big]$ are 
\begin{align}
\E_{\prior^\data} \E_\eta [ \| \mu_1^\data -\phi_{true} \|_{L^2_\rho}^2 ]
&= \sum_{1\leq i\leq K} (\lambda_i+\sigma_{\eta}^{2} \lambda_*\lambda_i^{-1} )^{-2} [\sigma_\eta^2  (\lambda_i +\sigma_{\eta}^{2} \lambda_*^2 \lambda_i^{-1} ) + |\epsilon^\xi_i  |^2], 
\label{eq:exp_MSE_MAP_data} \\
\E_{\prior} \E_\eta [ \| \mu_1 -\phi_{true} \|_{L^2_\rho}^2 ]
&=\sum_{1\leq i\leq K} (\lambda_i+\sigma_{\eta}^{2} r_i^{-1} )^{-2} [\sigma_\eta^2  \lambda_i +\sigma_{\eta}^{4} r_i^{-1}  + |\epsilon^\xi_i  |^2]+ \sum_{i>K}[r_i+\sigma_\eta^{-4} r_i^2 | \epsilon^\xi_i  | ^2]\,. \label{eq:exp_MSE_MAP_data2}
\end{align}
When $r_i=0$ for all $i> K$ and $\lambda_i = r_i$, $\lambda_*=1$ for all $i\leq K$, i.e., when the two priors are the same, the two expectations are equal. 

To prove \eqref{eq:MSEineq}, it suffices to compare the summations over $1\leq i\leq k$ in \eqref{eq:exp_MSE_MAP_data} and \eqref{eq:exp_MSE_MAP_data2}. Note that 
\begin{align*}
\lambda_*\leq  1 &\implies \lambda_i+\sigma_{\eta}^{2} \lambda_*^2\lambda_i^{-1} \leq \lambda_i+\sigma_{\eta}^{2} \lambda_*\lambda_i^{-1}, \\
\max_{1\leq i\leq K}\{\lambda_i r_i^{-1}\}\leq \lambda_* &\implies \lambda_i+\sigma_{\eta}^{2} \lambda_*\lambda_i^{-1}\geq \lambda_i+\sigma_{\eta}^{2} r_i^{-1}.
\end{align*}
Then, we have
\begin{align*}
\sum_{1\leq i\leq K} &\left(\lambda_i+\sigma_{\eta}^{2} \lambda_*\lambda_i^{-1}  \right)^{-2} [\sigma_\eta^2  (\lambda_i +\sigma_{\eta}^{2} \lambda_*^2 \lambda_i^{-1})  + |\epsilon^\xi_i  |^2] \\
&\leq \sum_{1\leq i\leq K} \left(\lambda_i+\sigma_{\eta}^{2} \lambda_*\lambda_i^{-1}  \right)^{-1} \sigma_\eta^2    + \left(\lambda_i+\sigma_{\eta}^{2} \lambda_*\lambda_i^{-1}  \right)^{-2}  |\epsilon^\xi_i  |^2 \\
&\leq \sum_{1\leq i\leq K} \left(\lambda_i+\sigma_{\eta}^{2} r_i^{-1}  \right)^{-1} \sigma_\eta^2    + \left(\lambda_i+\sigma_{\eta}^{2} r_i^{-1}  \right)^{-2}  |\epsilon^\xi_i  |^2 \\
& = \sum_{1\leq i\leq K} \left(\lambda_i+\sigma_{\eta}^{2} r_i^{-1}  \right)^{-2} [\sigma_\eta^2  \lambda_i +\sigma_{\eta}^{4} r_i^{-1}  + |\epsilon^\xi_i  |^2] \,.
\end{align*}
In particular, the first inequality is strict if $\lambda_*<  1$, and the second inequality is strict if $\lambda_i < r_i$ for some $1\leq i\leq K$. 
Thus, the inequality in \eqref{eq:MSEineq} is strict if the two priors differ. 
\end{proof}

Additionally, the next theorem
shows that under the condition $\lambda_*>\max_{i\leq K}\{\lambda_i r_i^{-1}\} $, the data-adaptive prior outperforms the non-degenerate prior in producing a posterior with a smaller trace of covariance. We note that this condition is sufficient but not necessary, since the proof is based on component-wise comparison and does not take into account the part $\sum_{i>K} r_i$ (see Remark \ref{rmk:spectrum_condition} for more discussions).  

\begin{theorem}[Trace of the posterior covariance] \label{thm:trace_post}
Suppose that Assumption {\rm \ref{assumption2} (A1-A2)} holds. Recall that $\calQ_1^\data$ and $\calQ_1$ are the posterior covariance operators of the data-adaptive prior and the non-degenerate prior in \eqref{eq:posterior_D} and \eqref{eq:posterior},  respectively.  Then, $Tr(\calQ_1^\data) < Tr(\calQ_1) $ if $\lambda_* >  \max_{i\leq K}\{\lambda_i r_i^{-1}\} $. Additionally, when $r_i= 0$ for all $i> K$, we have $Tr(\calQ_1^\data) > Tr(\calQ_1) $ if $\lambda_* < \min_{i\leq K}\{\lambda_i r_i^{-1}\} $. 
\end{theorem}
\begin{proof} By definition, the trace of the two operators are 
\begin{equation}\label{eq:trace_post}
\begin{aligned}
    Tr(\calQ_1^\data) &=\sum_{1\leq i\leq K} \sigma_\eta^2 (\lambda_i+\sigma_\eta^2\lambda_* \lambda_i^{-1})^{-1} , 
    \\
    Tr(\calQ_1) &=  \sum_{1\leq i\leq K} \sigma_\eta^2 (\lambda_i+\sigma_\eta^2 r_i^{-1})^{-1} + \sum_{i>K} r_i.  
\end{aligned}
\end{equation}
Thus, when $\lambda_*> \max_i\{\lambda_i r_i^{-1}\}$, we have $(\lambda_i+\sigma_\eta^2\lambda_* \lambda_i^{-1})^{-1}  < (\lambda_i+\sigma_\eta^2 r_i^{-1})^{-1}$ for each $i\geq K$, and hence  $Tr(\calQ_1^\data) < Tr(\calQ_1) $. The last claim follows similarly.  
\end{proof}

\begin{remark}[Expected MSE of the MAP and the trace of the posterior covariance]\label{rmk:trace_is_not_Emse}
When there is no model error, we have $ \E_{\pi_0}\E_\eta\Big[\|\mu_1-\phi_{true}\|_{L^2_\rho}^2\Big]= Tr(\calQ_1) $ in \eqref{eq:trace_post}. That is, for the prior $\pi_0$,  the expected MSE of the MAP estimator is the trace of the posterior covariance {\rm\cite[Theorem 2]{alexanderian2016bayesian}}. However, for the data-adaptive prior $\pi_0^\data$, we have $ \E_{\pi_0^\data} \E_\eta \Big[\|\mu_1^\data-\phi_{true}\|_{L^2_\rho}^2\Big] = Tr(\calQ_1^\data)$ \emph{if and only if} $\lambda_*=1$, which follows from \eqref{eq:exp_MSE_MAP_data} and \eqref{eq:trace_post}. Thus, if $\max_{i\leq K}\{\lambda_i r_i^{-1}\}\leq 1$, a smaller expected MSE of the MAP estimator in  Theorem {\rm\ref{thm:MSE}}  implies a smaller trace of the posterior covariance in Theorem {\rm\ref{thm:trace_post}}.  
\end{remark}

\begin{remark}[A-optimality]\label{rmk:Aoptimality}  
Theorem {\rm\ref{thm:trace_post}} shows that the data-adaptive prior achieves A-optimality among all priors with $\{r_i\}$ satisfying $\lambda_* >  \max_{i\leq K}\{\lambda_i r_i^{-1}\} $. Here, an A-optimal design is defined to be the one that minimizes the trace of the posterior covariance operator in a certain class ({\rm\cite{alexanderian2016bayesian} and \cite{chaloner1995bayesian}}). It is equivalent to minimizing the expected MSE of the MAP estimator (which is equal to $Tr(\calQ_1)$) through an optimal choice of the $\prior$. 
Thus, in our context, the A-optimal design seeks a prior with $\{r_i\}_{i\geq 1}$ in a certain class such that $g(r_1,\ldots,r_K) :=Tr(\calQ_1) = \sum_{i\leq K} (\lambda_i+r_i^{-1})$ is minimized, and the data-adaptive prior achieves A-optimality in the above class of priors. 
\end{remark}

\begin{remark}[Conditions on the spectra]\label{rmk:spectrum_condition}
The condition $\max_{i\leq K}\{\lambda_i r_i^{-1}\} \leq \lambda_*$ in Theorems {\rm \ref{thm:MSE}--\ref{thm:trace_post}} is far from necessary, since their proofs are based on a component-wise comparison in the sum and its does not take into account the part $\sum_{i>K} r_i$. 
 The optimal $\lambda_*$ in practice is often much smaller than the maximal ratio $\max_{i\leq K}\{\lambda_i r_i^{-1}\}$ and it depends on the dataset, in particular, it depends nonlinearly on all the elements involved (see Figures {\rm \ref{fig:lambda_stats}--\ref{fig:lambda_stats_inFSOI}} in Appendix {\rm \ref{appd-computeToeplitz}}). Thus, a full analysis with an optimal $\lambda_*$ is beyond the scope of this study. 
\end{remark}

\section{Computational practice}
\label{sec:num}
We have followed the wisdom of \cite{Stuart10} on ``\textit{avoid discretization until the last possible moment}'' so that we have presented the analysis of the distributions on $L^2_\rho$ using operators.  In the same spirit, we avoid selecting a basis for the function space until the last possible moment. The moment arrives now. Based on the abstract theory in the previous sections, we present the implementation of the data-adaptive prior in computational practice. We demonstrate it on Toeplitz matrices and integral operators, which represent finite- and infinite-dimensional function spaces of learning, respectively.  

In computational practice, the goal is to estimate the coefficient $c= (c_1,\ldots,c_l)^\top \in\R^{l\times 1}$ of $\phi= \sum_{i=1}^l c_i \phi_i$ in a prescribed hypothesis space $\mH = \mathrm{span}\{\phi_i\}_{i=1}^l \subset L^2_\rho$ with $l \leq \infty$, where the basis functions $\{\phi_i\}$ can be the B-splines, polynomials, or wavelets. Then, the prior and posterior are represented by distributions of $c\in \R^l$.  Note that the pre-specified basis $\{\phi_i\}$ is rarely orthonormal in $L^2_\rho$ because $\rho$ varies with data. Hence, we only require that the basis matrix 
\begin{equation}\label{eq:Bmat}
B =  [\innerp{\phi_i,\phi_j}_{L^2_\rho}]_{1\leq i,j\leq l} 
\end{equation}
is non-singular, i.e., the basis functions are linearly independent in $L^2_\rho$. This simple requirement reduces redundancy in basis functions.

In terms of $c$, the negative log-likelihood in \eqref{eq:lossFn2} for  $\phi= \sum_{i=1}^l c_i \phi_i$ reads  
 \begin{equation}\label{eq:lossFn_c}
\begin{aligned}
\calE(c)  &= \innerp{\LGbar \phi,\phi}_{L^2_\rho} -2 \innerp{\phi^\data,\phi}_{L^2_\rho}+C_N^f\\
&=c^\top \Abar  c- 2c^\top \bbar + C_N^f, 
\end{aligned}
\end{equation}
where the regression matrix $\Abar$ and vector $\bbar$ are given by 
\begin{subnumcases}{\label{eq:Abarbbar}}
	\Abar(i,j)  = \frac{1}{N}\sum_{1\leq k\leq N} \innerp{R_{\phi_i}[u^k], R_{\phi_j}[u^k]}_{\spaceY} = \innerp{\LGbar \phi_i,\phi_j}_{L^2_\rho}, \label{eq:Abarbbar_A}\\
\bbar(i)   =  \frac{1}{N}\sum_{1\leq k\leq N} \innerp{R_{\phi_i}[u^k], f^k}_{\spaceY} = \innerp{ \phi_i,\phi^\data}_{L^2_\rho}. \label{eq:Abarbbar_b}
\end{subnumcases}
The maximal likelihood estimator (MLE) $\widehat c = \Abar^{-1}\bbar$ is the default choice of solution when $\bbar$ is in the range of $\Abar$. 
However, the MLE is ill-defined when $\bbar$ is not in the range of $\Abar$, which may happen when there is model error (or computational error due to incomplete data, as we have discussed after Theorem \ref{thm:FSOI}), and a Bayesian approach makes the inverse problem well-posed by introducing a prior.

We will compare our data-adaptive prior with the widely-used standard Gaussian prior on \emph{the coefficient}, that is, $c\sim \pi_0= \calN(0,Q_0)$ with $Q_0= I$, the identity matrix on $\R^l$. This prior leads to a posterior  $\pi_1 = \calN(m_1,Q_1)$ with  
\begin{equation}
m_1= (\Abar + \sigma^2_\eta I)^{-1}\bbar, \quad Q_1 = (\Abar + \sigma^2_\eta I)^{-1}. 
\end{equation}

\subsection{Data-adaptive prior and posterior of the coefficient}
\label{sec:DAprior-comput}
The next proposition computes the prior and posterior distributions of the random coefficient $c=(c_1,\ldots,c_l)^\top \in\R^{l\times 1}$ of the $L^2_\rho$-valued random variable $\phi=\sum_{i} c_i \phi_i$ with the data-adaptive prior \eqref{eq:new_prior}. 

 \begin{proposition}\label{prop:prior_post_in_c} 
 Assume that  $ \{\phi_i\}_{i\geq 1}$ is a complete basis of $L^2_\rho$ that may not be orthonormal, and the basis matrix $B$ of $ \{\phi_i\}_{i=1}^l$ in \eqref{eq:Bmat} is invertible. Denote  $\phi=\sum_{i} c_i \phi_i$ the $L^2_\rho$-valued random variable with the data-adaptive prior in  \eqref{eq:new_prior}. 
 Then, the prior and posterior distributions of $c=(c_1,\ldots,c_l)^\top \in\R^{l\times 1}$ are $\calN(0,Q_0^\data)$ and $\calN(m_1^\data,Q_1^\data)$ with
 \begin{subequations}\label{eq:comput}
     \begin{align}
         Q_0^\data &= \lambda_*^{-1} B^{-1}\Abar B^{-1}\,, \label{eq:comput_0}\\
         Q_1^\data &= \sigma_\eta^2  (\Abar+ \sigma_\eta^2 \lambda_* B\Abar^{-1}B )^{-1}, \quad   m_1^\data = \sigma_\eta^{-2}Q_1^\data \bbar,  \label{eq:comput_1}
     \end{align}
 \end{subequations}
 where $\Abar$ and $\bbar$ be defined in \eqref{eq:Abarbbar}. 
 \end{proposition}
 \begin{proof} The prior covariance \eqref{eq:comput_0} follows directly from the definition of the data-adaptive prior in \eqref{eq:new_prior} and Lemma \ref{lemma:Gausian_basis}. The posterior covariance and mean follow from the likelihood in \eqref{eq:lossFn_c} and the $ Q_0^\data$ above: 
 $$ 
 \frac{d\posterior^\data(c) }{dc}\propto \exp\left(-\frac{1}{2}\big[ \sigma_{\eta}^{-2} (c^\top \Abar c- 2c^\top \bbar + C_N^f)+ c^\top(Q_0^{\data})^{-1}c \big] \right).
 $$ 
 Thus, completing the squares in the exponent, we obtain \eqref{eq:comput_1}. 
 \end{proof}


\begin{remark}[Relation between distributions of the coefficient and the function] \label{rmk:coef_operator}
The prior and posterior distributions of the coefficient $c$ and the function $\phi= \sum_{i=1}^l c_i\phi_i$ are different: the former depends on the basis $\{\phi_i\}_{i=1}^l$, but the latter is not. The relation between the distributions of $c$ and $\phi$ is characterized by Lemma {\rm\ref{lemma:Gausian_basis}--\ref{lemma:basis_to_operator}}. Specifically, if $c\sim \calN(0,Q)$  and $\phi= \sum_{i=1}^l c_i\phi_i$ has a Gaussian measure $\calN(0,\calQ)$ on $\mH=\mathrm{span}\{\phi_i\}_{i=1}^l$, then, we have $A:=(\innerp{\phi_i,\calQ \phi_j}) =BQB$ provided that $B$ in \eqref{eq:Bmat} is strictly positive definite. Additionally, when computing the trace of the operator $\calQ$, we solve a generalized eigenvalue problem $Av = \lambda Bv$, which follows from the proof of Proposition {\rm\ref{lemma:LG-A}} below. 
\end{remark}

\begin{remark}[Relation to the basis matrix of the RKHS]\label{rmk:relation_rkhs}
 The matrix $B^{-1}\Abar B^{-1}$ in the covariance $Q_0^\data$ in \eqref{eq:comput_0} is the pseudo-inverse of the basis matrix of $\{\phi_i\}$ in the RKHS $H_G$ defined in Lemma {\rm \ref{lemma:sidaRKHS}}, that is, $B_{rkhs}(i,j) =\innerp{\phi_i,\LGbar^{-1} \phi_j}_{L^2_\rho}= \innerp{\phi_i,\phi_j}_{H_G}$, assuming that the basis functions $\{\phi_i\}$ are in the RKHS. Computation of the matrix $B_{rkhs}$ involves a general eigenvalue problem to solve the eigenvalues of $\LGbar$ (see Proposition {\rm\ref{lemma:LG-A}}).  
\end{remark}

We select the hyper-parameter $\lambda_*$ by the L-curve method in \cite{hansen_LcurveIts_a}. The L-curve is a log-log plot of the curve 
$ l(\lambda)=(y(\lambda),x(\lambda))$ with $y(\lambda) ^2= c_\lambda^\top B\Abar^{-1} Bc_\lambda $ and $x(\lambda)^2 = \calE(c_\lambda) $, where $c_\lambda = (\Abar+\lambda B\Abar^{-1} B )^{-1} \bbar$. 
The L-curve method maximizes the curvature to balance between the minimization of the likelihood and the control of the regularization:  
$$ 
	\lambda_{*} 
	= \underset{{\lambda_{\text{min}} \leq \lambda \leq \lambda_{\text{max}}}}{\rm{argmax}} \kappa(\it{l} (\lambda)),   \quad \kappa(\it{l} (\lambda))= \frac{x'y'' - x' y''}{(x'\,^2 + y'\,^2)^{3/2}},
$$ 
where $\lambda_{\min}$ and $\lambda_{\max}$ are the smallest and the largest generalized eigenvalues of $(\Abar,B)$.  

We summarize the priors and posteriors in computation in Table \ref{tab:priors_posteriors_compute}.
\begin{table}[htp] 
	\begin{center} 
		\caption{  \text{Priors and posteriors of the coefficients $c$ of $\phi = \sum_{i=1}^l c_i \phi_i \in  \mH\subset L^2_\rho$}. 
		 } \label{tab:priors_posteriors_compute}
		\begin{tabular}{ l   l  l }		\toprule 
                 Gaussian measure                           & Mean  & Covariance  \\  \hline
	         $\prior  = \calN(m_0,Q_0)$   &   $m_0=0$  & $Q_0=I$\\ 
	         $\posterior = \calN(m_1,Q_1)$      &   $m_1 = (  \Abar+ \sigma_{\eta}^{2} I )^{-1} \bbar $  & $Q_1 = \sigma_{\eta}^{2} (  \Abar + \sigma_{\eta}^{2} I )^{-1}  $ \vspace{2mm} \\
	           \hline  
	         $\prior^\data =\calN(m_0^\data,Q_0^\data)$   &   $m_0^\data=0$  & $Q_0^\data = \lambda_*^{-1} B^{-1}\Abar B^{-1} $ \rule{0pt}{2.5ex}\\ 
	        $\posterior^\data = \calN(m_1^\data,Q_1^\data)$      &   $m_1^\data=   \sigma_\eta^{-2}   Q_1^\data \bbar$  & $Q_1^\data =  \sigma_\eta^2  (\Abar + \sigma_\eta^2 \lambda_* B \Abar^{-1}B  )^{-1}$ \\
  \hline
	 			\bottomrule	  
		\end{tabular}  \vspace{-4mm}
	\end{center}
\end{table}

\begin{remark}[Avoiding pseudo-inverse of singular matrix]\label{rmk:avoid_pinv}
The inverse of the matrix in $Q_1^\data$ in \eqref{eq:comput_1} can cause a large numerical error when $\Abar$ is singular or severely ill-conditioned. We increase the numerical stability by avoiding  $\Abar^{-1}$: let $D= B^{-1}\Abar^{1/2} $ and write $Q_1^\data$ as 
\begin{equation}\label{eq:Q1data_comput}
Q_1^\data = \sigma_\eta^2  (\Abar+ \sigma_\eta^2 \lambda_* B\Abar^{-1}B )^{-1} = \sigma_\eta^2 D (D^{\top} \Abar D  +\lambda I)^{-1} D^\top. 
\end{equation}
\end{remark}

\begin{remark}[Relation to Zellner's g-prior]\label{rmk:Zellner} When the basis of the hypothesis space are orthonormal in $L^2_\rho$ (that is, the basis matrix $B= (\innerp{\phi_i,\phi_j}_{L^2_\rho})_{1\leq i,j\leq l}= I$), we have $Q_0^{\data} = \Abar$. Thus, we are once again revealing the well-known Zellner's g-prior {\rm\cite{Agliari1988gprior,Bayarri2012Criteria,Zellner1980gprior}}. \end{remark}

The next proposition shows that the eigenvalues of $\LGbar$ are solved by a generalized eigenvalue problem. Its proof is deferred to Appendix \ref{append_ID}. 
\begin{proposition} \label{lemma:LG-A} 
Assume that the hypothesis space satisfies $\mH = \mathrm{span}\{\phi_i\}_{i=1}^l \supseteq \LGbar(L^2_\rho) $ with $l\leq \infty$, where $\LGbar:L^2_\rho\to L^2_\rho$ be the integral operator in \eqref{eq:LG}. Let $\Abar$ and $\bbar$ be defined in \eqref{eq:Abarbbar}.  
Then, the operator $\LGbar$ has eigenvalues $(\lambda_1,\ldots,\lambda_l)$ solved by the generalize eigenvalue problem with $B$ in \eqref{eq:Bmat}: 
\begin{equation}\label{eq:gEigenP}
\Abar V=  B  V\Lambda, \quad s.t., V^\top B V = I, \quad \Lambda= \mathrm{Diag}(\lambda_1,\ldots,\lambda_l). 
\end{equation}
and the corresponding eigenfunctions of $\LGbar$ are  $\{\psi_k = \sum_{j=1}^l  V_{jk}\phi_j\}$. Additionally, for any $\phi=\sum_{i}^{l}c_i\phi_i $ in $\LGbar^{1/2}(L^2_\rho)$, we have $\innerp{\phi,\LGbar^{-1}\phi}_{L^2_\rho} = c^\top B_{rkhs} c  $ with 
\[
B_{rkhs}  =  (V\Lambda V^\top)^{-1} = B\Abar^{-1} B.
\]
\end{proposition}


\subsection{Example: discrete kernels in Toeplitz matrices}\label{sec:Toeplitz}
The Toeplitz matrix in Example \ref{exp:Toeplitz_matrix} has a vector kernel, which lies in a finite-dimensional function space of learning $L^2_\rho$. It provides a typical example of discrete kernels. We use the simplest case of a $2\times 2$ Toeplitz matrix to demonstrate the data-adaptive function space of identifiability and the advantages of the data-adaptive prior.  

We aim to recover the kernel  $\phi\in \R^{2n-1}$ in the $\R^{n\times n}$ Toeplitz matrix from measurement data $\{(u^k,f^k)\in \R^n\times \R^n\}_{k=1}^N$ by fitting the data to the model \eqref{eq:toeplitz}. 
The kernel is a vector $\phi: \calS\to \R^{2n-1}$ with $\calS= \{r_l\}_{l=1}^{2n-1}$ with $r_l= l-n$. Since $R_\phi[u]$ is linear in $\phi$, there is a matrix $L_u\in \R^{n\times (2n-1)}$ such that $R_\phi [u] = L_u \phi$. Note that $L_u$ is linear in $u$ since $R_\phi [u]$ is, hence only linearly independent data $\{u^k\}_{k+1}^N$ brings new information for the recovery of $\phi$.

A least squares estimator (LSE) of  $\phi\in \R^{2n-1}$ is 
 \[ \widehat \phi = \Abar^{-1}\bbar, \,  \text{ with  } \Abar = \frac{1}{N}\sum_{1\leq k\leq N}L_{u^k}^\top L_{u^k},\quad  \bbar =   \frac{1}{N}\sum_{1\leq k\leq N}L_{u^k}^\top f^k, \] 
Here the $\Abar^{-1}$ is a pseudo-inverse when $\Abar$ is singular. 
However, the pseudo-inverse is unstable to perturbations, and the inverse problem is ill-posed.

We only need to identify the basis matrix $B$ in \eqref{eq:Bmat} to get the data-adaptive prior and its posterior in Table \ref{tab:priors_posteriors_compute}. The basis matrix requires two elements: the exploration measure and the basis functions. Here the exploration measure $\rho$ in \eqref{eq:rho_conti} is 
 $\rho(r_l) =Z^{-1}\sum_{1\leq k\leq N} \sum_{0\leq i,j\leq n}  \delta(i-j-r_l) |u_j^k|
$ with $r_l\in \calS$, 
where $Z = n\sum_{k=1}^N\sum_{i=1}^{n-1} |u_i^k| $ is the normalizing constant. 
Meanwhile, the unspoken hypothesis space for the above vector $\phi = \sum_{i=1}^{2n-1} c_i \phi_i$ with $c_i= \phi(r_i)$ is $\mH = \mathrm{span}\{\phi_i\}_{i=1}^{2n-1} = \R^{2n-1}$ with basis $\phi_i(r)=\delta(r_i-r)\in L^2(\calS, \R)$, where $\delta$ is the Kronecker delta function. Then, the basis matrix of $\{\phi_i(r)=\delta(r_i-r)\}$ in $L^2_\rho$, as defined in \eqref{eq:Bmat}, is $ B= \mathrm{Diag}(\rho)$. 
Thus, if $\rho$ is not strictly positive, this basis matrix is singular and these basis functions are linearly dependent (hence redundant) in $L^2_\rho$. In such a case, we select a linearly independent basis for $L^2_\rho$, which is a proper subspace of $\R^{2n-1}$, and we use pseudo-inverse of $\Abar$ and $B$ to remove the redundant rows. Additionally, since vector $\phi$ is the same as its coefficient $c$, the priors and posteriors in Table \ref{tab:priors_posteriors} and Table \ref{tab:priors_posteriors_compute} are the same.

\paragraph{Toeplitz matrix with $n=2$.} 
Table  \ref{tab:Toeplitz_rho} shows three representative datasets for the inverse problem:  (1) the dataset $\{u^1= (1,0)\}$ leads to a well-posed inverse problem in $L^2_\rho$ though it appears ill-posed in $\R^3$, (2) the dataset $\{u^1,u^2=(0,1)\}$ leads to a well-posed inverse problem, and (3) the dataset $\{u^3 = (1,1)\}$ leads to an ill-posed inverse problem and our data-adaptive prior significantly improves the accuracy of the posterior, see Table \ref{tab:Toeplitz_map}. Computational details are presented in Appendix \ref{appd-computeToeplitz}.

\begin{table}[h!]
\centering
\caption{The exploration measure, the FSOI and the eigenvalues of $\LGbar$ for learning the kernel in a $2\times 2$ Toeplitz matrix from 3 typical datasets. } \label{tab:Toeplitz_rho}
\vspace{-2mm}
\begin{tabular}{cc cccc}
\toprule\hline 
Data $\{u^k\}$            & $\rho$ on $\{-1,0,1\}$                                & FOSI            & Eigenvalues of $\LGbar$  \rule[-1.2ex]{0pt}{0pt}  \\   \hline 
$\{ u^1= (1,0)^\top\}$    &  $ (0,\frac{1}{2},\frac{1}{2}) $     & $\mathrm{span}\{\phi_2,\phi_3\} = L^2_\rho$     & $\{1,1\}$ \rule{0pt}{2.5ex}\\  
$\{u^1,u^2=(0,1)\}$         &    $ (\frac{1}{4},\frac{1}{2},\frac{1}{4}) $  & $\mathrm{span}\{\phi_1,\phi_2,\phi_3\}= L^2_\rho$  & $\{2,2,2\}$ \rule{0pt}{2.5ex} \\  
$\{u^3=(1,1)^\top\}$     &  $ (\frac{1}{4},\frac{1}{2},\frac{1}{4}) $  & $\mathrm{span}\{\psi_1,\psi_2\}\subsetneq L^2_\rho$ & $\{8,4,0\}$ \rule{0pt}{2.5ex} \\ 
 \hline\bottomrule
\end{tabular}
\caption*{{\small $^*$The basis $\{\phi_i\}$ are defined as $(\phi_1,\phi_2,\phi_3)=I_3$. For the dataset $\{u^ 3\}$, the eigenvectors of $\LGbar$ in $L^2_\rho$ are $\psi_1 = (1,1,1)^\top$, $\psi_2 = (-\sqrt{2},0,\sqrt{2})^\top,$ and $ \psi_3 = (1,-1,1)^\top$, see the text for more details.  
}}\vspace{-3mm}
\end{table}

\begin{table}[h!]
\centering
\caption{Performance of the posteriors in learning the kernel of Teoplitz matrix.$^*$  } \label{tab:Toeplitz_map} \vspace{-2mm}
\begin{tabular}{c cc  | c c} \toprule\hline 
 $\phi_{true}$   & Bias of $m_1$   & Bias of $m_1^\data$             & $Tr(\calQ_1)$   & $Tr(\calQ_1^\data)$    \vspace{1mm}  \\   \hline \rule{0pt}{1\normalbaselineskip}
 $(1,1,1)^\top\in $ FSOI    & $0.34 \pm 0.01 $    &  $\mbf{ 0.10 \pm 0.11}$          &  $0.34 \pm 0.00$  & $\mbf{ 0.0037 \pm 0.00 }$ \vspace{1mm} \\ 
 $(1,0,1)^\top\notin $ FSOI   & $0.94 \pm 0.01$  &  $\mbf{ 0.66 \pm 0.09}$           &  $0.34 \pm 0.00$  & $\mbf{ 0.0037 \pm 0.00 }$  \\ \hline \bottomrule
\end{tabular}

\caption*{{\footnotesize* We compute the means and standard deviations of the relative errors of the posterior means (``bias of $m_1$'' and ``bias of $m_1^\data$'') and the traces of the covariance of posteriors. They are computed in 100 independent datasets with $f^{3}$ observed with random noises, which are sampled from $\calN(0,\sigma_\eta^2)$ with $\sigma_\eta=0.1$. and the $u$ data is $\{u^3=(1,1)\}$. 
The relative bias of each estimator $m$ is computed by $\|m-\phi_{true}\|_{L^2_\rho}/ \|\phi_{true}\|_{L^2_\rho}$. The standard deviations of the traces are less than $10^{-5}$. 
}} \vspace{-3mm} 
\end{table}

Table \ref{tab:Toeplitz_map} demonstrates the significant advantage of the data-adaptive prior over the non-degenerate prior in the case of the third dataset. 
We examine the performance of the posterior in two aspects: the trace of its covariance operator, and the bias in the posterior mean. Following Remark \ref{rmk:coef_operator}, we compute the trace of the covariance operator of the posterior by solving a generalized eigenvalue problem. Table \ref{tab:Toeplitz_map} presents the means and standard deviations of the traces and the relative errors of the posterior mean. It consider two cases: $\phi_{true}=\psi_1$ in the FSOI and $\phi_{true}=(1,0,1)^\top=0.5\psi_1+0.5\psi_3$ outside of the FSOI (see Table \ref{tab:Toeplitz_rho}). 
We highlight two observations. 
\begin{itemize}[leftmargin=*]
\item The posterior mean $m_1^\data$ is much more accurate than $m_1$. When $\phi_{true}$ is in the FSOI, $m_1^\data$ is relatively accurate. When $\phi_{true}$ is outside the FSOI, the major bias comes from the part outside the FSOI, i.e., the part $0.5\psi_3$ leads to a relatively large error. 
\item The trace of $\calQ_1^\data$ is significantly smaller than $\calQ_1$. Here $\calQ_1^\data$ has a zero eigenvalue in the direction outside of the FSOI, while $\calQ_1$ is full rank. 
\end{itemize}

\paragraph{Discrete inverse problem: stable small noise limit.}

For the discrete inverse problem of solving $\phi \in \R^m$ in $L_k \phi = f_k$ for $1\leq k\leq N$, the next proposition shows that Assumption \ref{assumption2} (A3) does not hold, regardless of the presence of model error or missing data in $f_k$. Thus, it has a small noise limit. 
Numerical tests confirm that $\bbar $ is always in the range of the operator $\Abar$ and that $m_1$ has a small noise limit regardless of the model error (e.g., $\xi(u)= 0.01 u|u|^2$) or computational error due to missing data. 
However, for continuous inverse problems that estimate a continuous function $\phi$, Assumption \ref{assumption2} (A3) holds when $\bbar$ is computed using different regression arrays from those in $\Abar$ due to discretization or missing data (see Sect.\ref{sec:integral-operator}) or errors in integration by parts \cite{LangLu22}. 
\begin{proposition}\label{prop:lse-in-FSOI}
Let $\Abar = \sum_{1\leq k\leq N} L_k^\top L_k$ and $\bbar =  \sum_{1\leq k\leq N}  L_k^\top f_k$, where $L_k\in \R^{n\times m}$ and $f_k\in \R^{n\times 1}$ for each $1\leq k\leq N$. Then, $\bbar\in \text{Range}(\Abar) $. 
\end{proposition}
\begin{proof}
First, we show that it suffices to consider $L_k$'s being rank-1 arrays. The SVD (singular value decomposition) of each $L_k$ gives $L_k= \sum_{1\leq i\leq n_k} \sigma_{k,i} w_{k,i} v_{k,i}^\top$, where $\{ \sigma_{k,i}, w_{k,i}, v_{k,i}\}$ are the singular values, left and right singular vectors that are orthonormal, i.e., $w_{k,i}^\top w_{k,j} =\delta_{i,j}$ and $v_{k,i}^\top v_{k,j} =\delta_{i,j}$.  Denote $L_{k,i}=  \sigma_{k,i} w_{k,i} v_{k,i}^\top$, which is rank-1. Note that  $L_k^\top L_k = \sum_{1\leq i,j\leq n_k}   \sigma_{k,i}^2 v_{k,i} w_{k,i}^\top w_{k,j} v_{k,j}^\top = \sum_{1\leq i \leq n_k}   \sigma_{k,i}^2 v_{k,i} v_{k,i}^\top =  \sum_{1\leq i \leq n_k} L_{k,i}^\top L_{k,i} $. Thus,  we can write $\Abar= \sum_{1\leq k\leq N}\sum_{1\leq i \leq n_k} L_{k,i}^\top L_{k,i}$ and $\bbar = \sum_{1\leq k\leq N}\sum_{1\leq i \leq n_k} L_{k,i}^\top f_{k}$ in terms of rank-1 arrays. 

Next, for each $k$, write the rank-1 array as $L_k= \sigma_k w_k v_k^\top$ with $w_k\in \R^{m\times 1}$ and $v_k\in \R^{n\times 1}$ both being unitary vectors. Then, $\Abar = \sum_{1\leq k\leq N} \sigma_k^2 v_k w_k^\top w_k v_k^\top =  \sum_{1\leq k\leq N} \sigma_k^2 v_k v_k^\top$, and $\text{Range}(\Abar) = \mathrm{span}\{v_k\}_{k=1}^N$ (where the $v_k$'s can be linearly dependent).  Therefore, $\bbar =  \sum_{1\leq k\leq N} \sigma_k v_k v_k^\top f_k  $ is in the range of $\Abar$ because $v_k^\top f_k$ is a scalar. 
\end{proof}

\subsection{Example: continuous kernels in integral operators}\label{sec:integral-operator}
For the continuous kernels of the integral operators in Examples \ref{exp:integral}-\ref{exp:interaction}, their function space of learning $L^2_\rho$ is infinite-dimensional. Their Bayesian inversions are similar, so we demonstrate the computation using the convolution operator in Example \ref{exp:integral}. In particular, we compare our data-adaptive prior with a fixed non-degenerate prior in the presence of four types of errors: (i) discretization error, (ii) model error, (ii) partial observation (or missing data), and (iv) wrong noise assumption. 

Recall that with $\spaceX=\spaceY=L^2([0,1])$, we aim to recover the kernel $\phi$ in the operator in \eqref{eq:Int_operator}, 
 $   R_\phi[u](y) = \int_{0}^1 \phi(y-x) u(x)dx$, 
by fitting the model \eqref{eq:map_R} to an input-output dataset $\left\{ u^k, f^k \right\}_{k = 1}^3$. We set $\{u^k\}_{k = 1}^{3}$ to be the probability densities of normal distributions $\mN(-1.6+0.6k, 1/15)$ for $k=1,2,3$ and we compute $R_{\psi}[u^k] = \int_0^1 \psi(y-x)u^k(x)dx$ by the global adaptive quadrature method in \cite{shampine2008vectorized}. The data are $\left\{ u^k(x_j), f^k(y_l) \right\}_{k = 1}^3$ on uniform meshes $\{x_{j}\}_{j= 1}^{J}$ and $\{y_l\}_{l = 1}^L$ of $[0, 1]$ with $J=100$ and $L=50$. Here $f^k(y_l)$ is generated by
\begin{equation}\label{eq:fk-comput}
    f^k(y_l) = R_\phi[u^k](y_l) + \eta_l^k + \xi^k(y_l), 
\end{equation}
where $\eta_l^k$ are i.i.d.~$\calN(0,\sigma_\eta^2)$ random variables (unless the wrong noise assumption case to be specified later) with variance $\frac{\sigma_\eta^2}{\Delta y}$   
 and $\xi^k(y) = \sigma_\xi u(y)\abs{u(y)}$ are artificial model errors with $\sigma_{\xi}=0$ (no model error) or $\sigma_{\xi}=0.01$ (a small model error). 
\begin{figure}[ht]
\caption{The exploration   measure and the eigenvalues of the basis matrix $B$, regression matrix $A_D$ and operator $\LGbar$ (computed via the generalized eigenvalue problem of $(A_D,B)$). }\vspace{-2mm}
\centering
 \includegraphics[width=\textwidth]{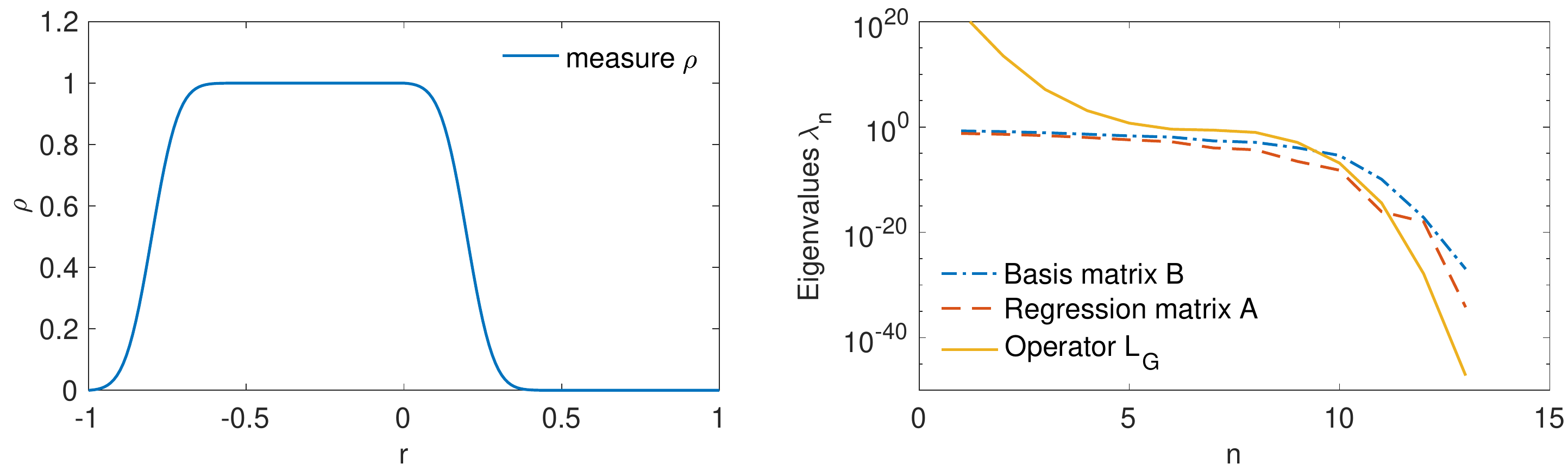} \vspace{-4mm}
 \label{fig:rho-eigenvalues} \ifjournal \vspace{-3mm} \fi 
\end{figure}

The exploration measure (defined in \eqref{eq:rho_conti}) of this dataset has a density 
  $$\rho(r) =   \frac{1}{ZN}\sum_{k = 1}^N \int_{[0,1]\cap[r, r+1]}  \abs{u^k(y)}dy, \quad r\in [-1, 1], $$  with $Z$ being the normalizing constant. We set the $\mH = \mathrm{span}\{\phi_i\}_{i=1}^l$, where $\{\phi_i\}_{i=1}^{l}$ are B-spline basis functions (i.e., piecewise polynomials) with degree 3 and with knots from a uniform partition of $[-1,1]$.  
We approximate $\Abar$ and $\bbar$ using the Riemann sum integration,  
\begin{equation*}
\begin{aligned}
    \Abar_D(i, i') & = \frac{1}{N}\sum_{k=1}^N\sum_{j = 1}^J \widehat{R}_{\phi_i}[u^k](y_j)\widehat{R}_{\phi_{i'}}[u^k](y_j)\Delta y,  \\
            \bbar(i) &=  \frac{1}{N}\sum_{1\leq k\leq N}\sum_{l = 1}^L \widehat{R}_{\phi_i}[u^k](x_l)f^k(y_l)  \Delta y,
        \end{aligned}
\end{equation*}
where we approximate $R_{\psi}[u^k]$ via Riemann integration  
$    \widehat{R}_{\psi}[u^k](y) = \sum_{j = 1}^J \psi(y - x_j) u^k(x_j) \Delta x
$. 
Additionally, to illustrate the effects of discretization error, we also compute $\Abar$ in \eqref{eq:Abarbbar} using the continuous $\{u^k\}$ with quadrature integration, and denote the matrix by $\Abar_C$. 

Figure \ref{fig:rho-eigenvalues} shows the exploration measure and the eigenvalues of the basis matrix $B$, $A_D$ and $\LGbar$ (which are the generalized eigenvalues of $(\Abar_D,B)$). Note that the support $\mS$ is a proper subset of $[-1,1]$, leading to a near singular $B$. In particular, the inverse problem is severely ill-posed in $L^2_\rho$ since $\LGbar$ has multiple almost-zero eigenvalues.  

We consider four types of errors, in addition to the observation noise, in $\bbar$ that often happen in practice. 
\begin{enumerate} 
    \item Discretization Error. We assume that $f^k$ in \eqref{eq:fk-comput} has no model error. 
    \item Partial Observation. We assume that $f^k$ misses data in the first quarter of the interval, i.e. $f^k_l = 0$ for $l = 0, \dots, L/4$. Also, assume that there is no model error. 
    \item Model Error. Assume there are model errors. 
    \item Wrong Noise Assumption. Suppose that the measurement noise $\eta_l^k$ is uniformly distributed on the interval $[-\frac{\sqrt{3}\sigma_\eta}{\sqrt{\Delta y}}, \frac{\sqrt{3}\sigma_\eta}{\sqrt{\Delta y}}]$. Thus, the model, which assumes a Gaussian noise, has a wrong noise assumption. Notice that we add a $\sqrt{3}$ to keep the variance at the same level as the Gaussian. 
\end{enumerate}
For each of the four cases, we compute the posterior means in Table \ref{tab:priors_posteriors_compute}  with the optimal hyper-parameter $\lambda_*$ selected by the L-curve method, and report the $L^2_\rho$ error of the function estimators. Additionally, for each of them, we consider different levels of observation noise $\sigma_\eta$ in $10^{-1} \sim 10^{-5}$, so as to demonstrate the small noise limit of the posterior mean.

\vspace{2mm}
We access the performance of the fixed prior and the data-adaptive prior in Table \ref{tab:priors_posteriors_compute} through the accuracy of their posterior means. We report the interquartile range (IQR, the $75^{th}$, $50^{th}$ and $25^{th}$ percentiles) of the $L^2_\rho$ errors of their posterior means in $200$ independent simulations in which $\phi_{true}$ are randomly sampled.

Two scenarios are considered:  $\phi_{true}$ is either inside or outside the FSOI.  To draw samples of $\phi_{true}$ outside the FSOI, we sample the coefficient $c^*$ of $\phi_{true}=\sum_{j=1}^{l} c_j^* \phi_j$ from the fixed prior $\calN(0,I_{l})$. Thus, the fixed prior is the true prior. To sample $\phi_{true}$ inside the FSOI, we sample $\phi_{true}=\sum_{j=1}^{l} c_j^* \psi_j$ with $c_*$ from $\calN(0,I_{3})$, where $\{\psi_j= \sum_{i=1}^l v_{i,j} \phi_j \}$ has $v_{\cdot,j}$ being the $j$-th eigenvector of $\Abar_D$. 

Note that the exploration measure, the matrices $\Abar_D$, $\Abar_C$ and $B$ are the same in all these simulations because they are determined by the data $\{u^k\}$ and the basis functions. Thus, we only need to compute $\bbar$ for each simulation.  

\begin{figure}[h!]
\caption{Interquartile range (IQR, the $75^{th}$, $50^{th}$ and $25^{th}$ percentiles) of the $L^2_\rho$ errors of the posterior means. They are computed in $200$ independent simulations with $\phi_{true}$ sampled from the fixed prior (hence {\bf outside the FSOI}), in the presence of four types of errors. 
 Top row: the regression matrix $\Abar$ is computed from continuous $\{u^k\}$; Bottom row: $\Abar$ is computed from discrete data. As $\sigma_\eta\to 0$, the fixed prior leads to diverging posterior means in 6 out of the 8 cases, while the data-adaptive (DA) prior leads to stable posterior means. 
} \label{fig:int_opt} \vspace{-2mm}
\centering
 \includegraphics[width=1\textwidth]{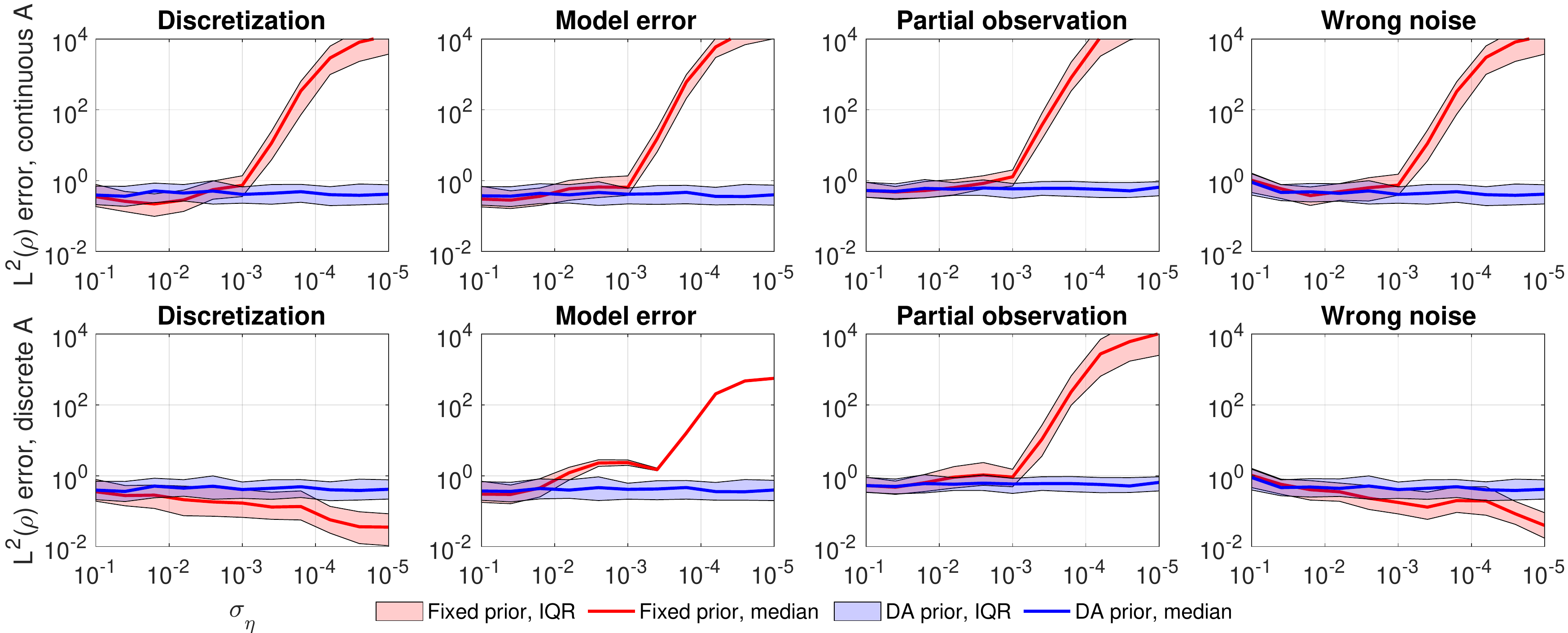} \vspace{-3mm}
\end{figure}

\begin{figure}[t!]
\caption{IQR of the $L^2_\rho$ errors of the posterior means in $200$ independent simulations with $\phi_{true}$ sampled {\bf inside the FSOI}. 
}\vspace{-2mm}
\centering
 \includegraphics[width=1\textwidth]{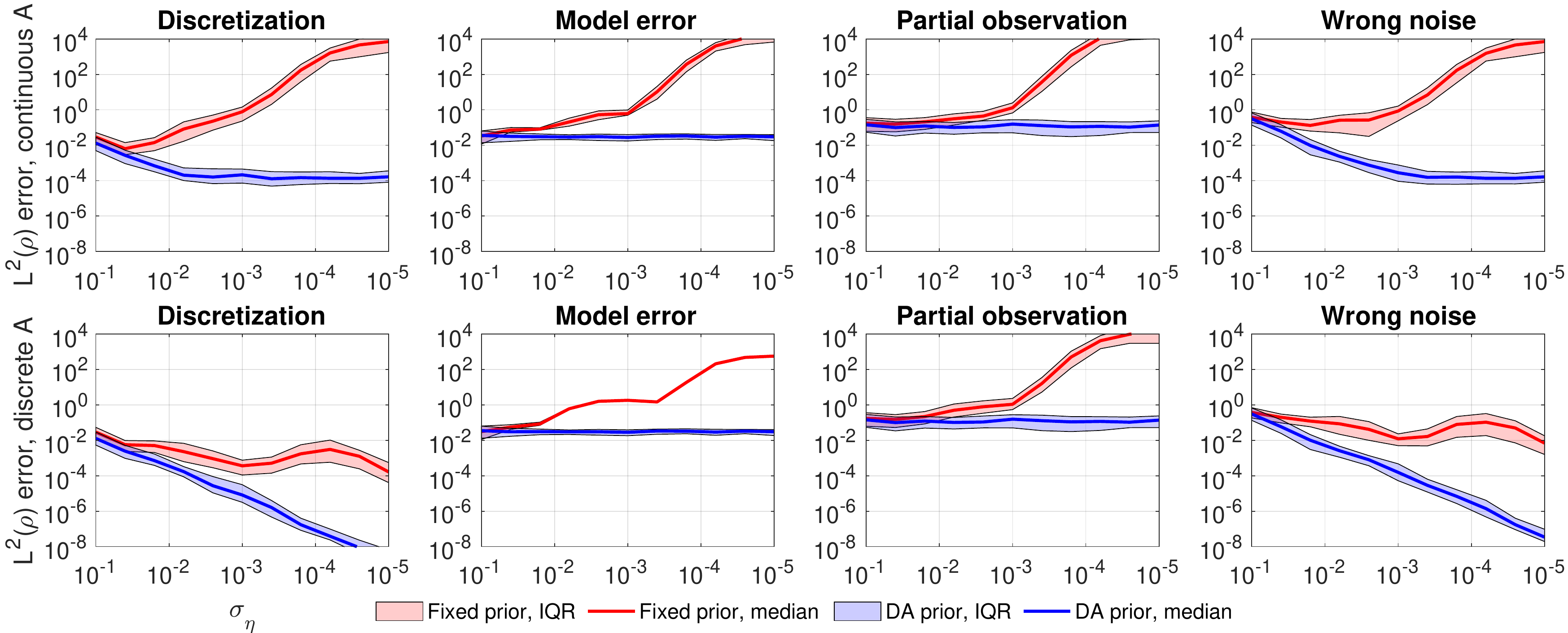}\vspace{-3mm}
\label{fig:int_opt_in_FSOI}
\ifjournal \vspace{-3mm} \fi 
\end{figure}

\begin{figure}[h!]
\caption{ The posterior (its mean, the $75^{th}$ and $25^{th}$ percentiles) when $\phi_{true} \notin$ FSOI.  
}\vspace{-2mm}
\centering
 \includegraphics[width=1\textwidth]{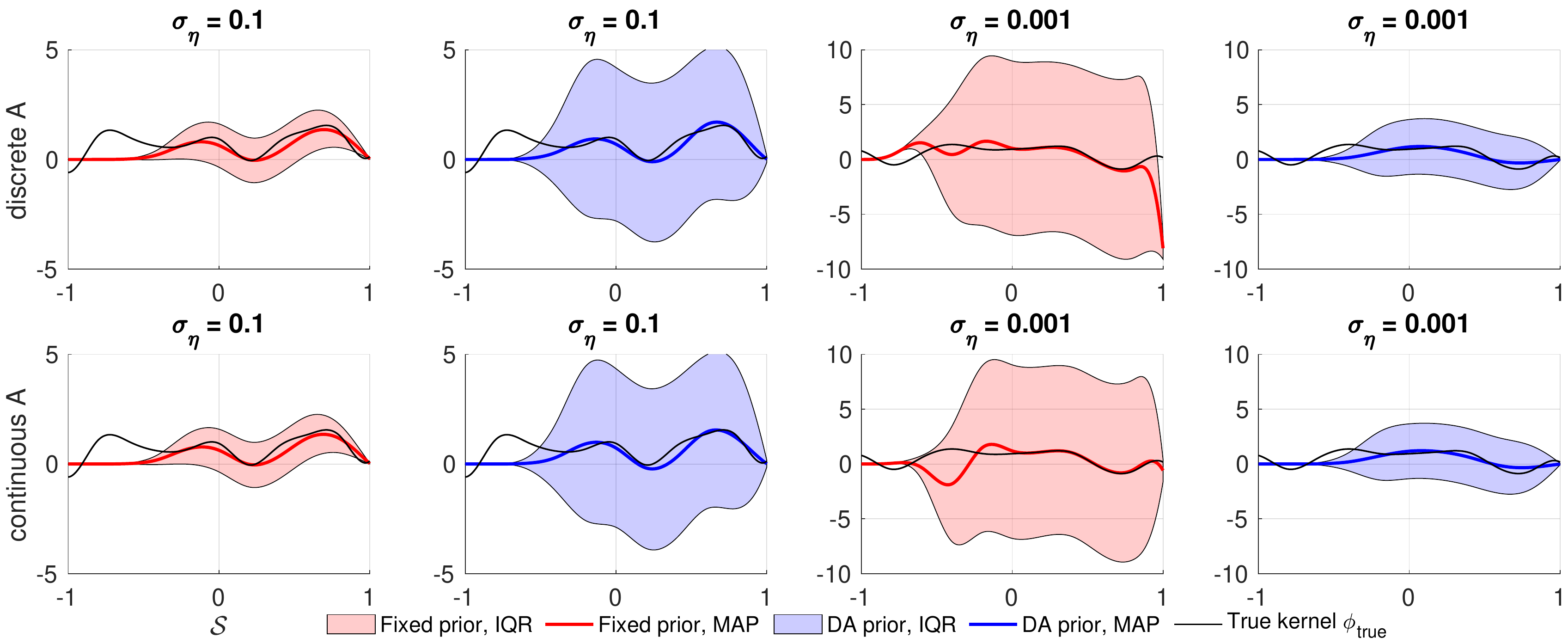}\vspace{-3mm}
\label{fig:post_outside_FSOI}
\ifjournal \vspace{-3mm} \fi 
\end{figure}

\begin{figure}[h!]
\caption{The posterior (its mean, the $75^{th}$ and $25^{th}$ percentiles) when $\phi_{true} \in$ FSOI. 
}\vspace{-2mm}
\centering
 \includegraphics[width=1\textwidth]{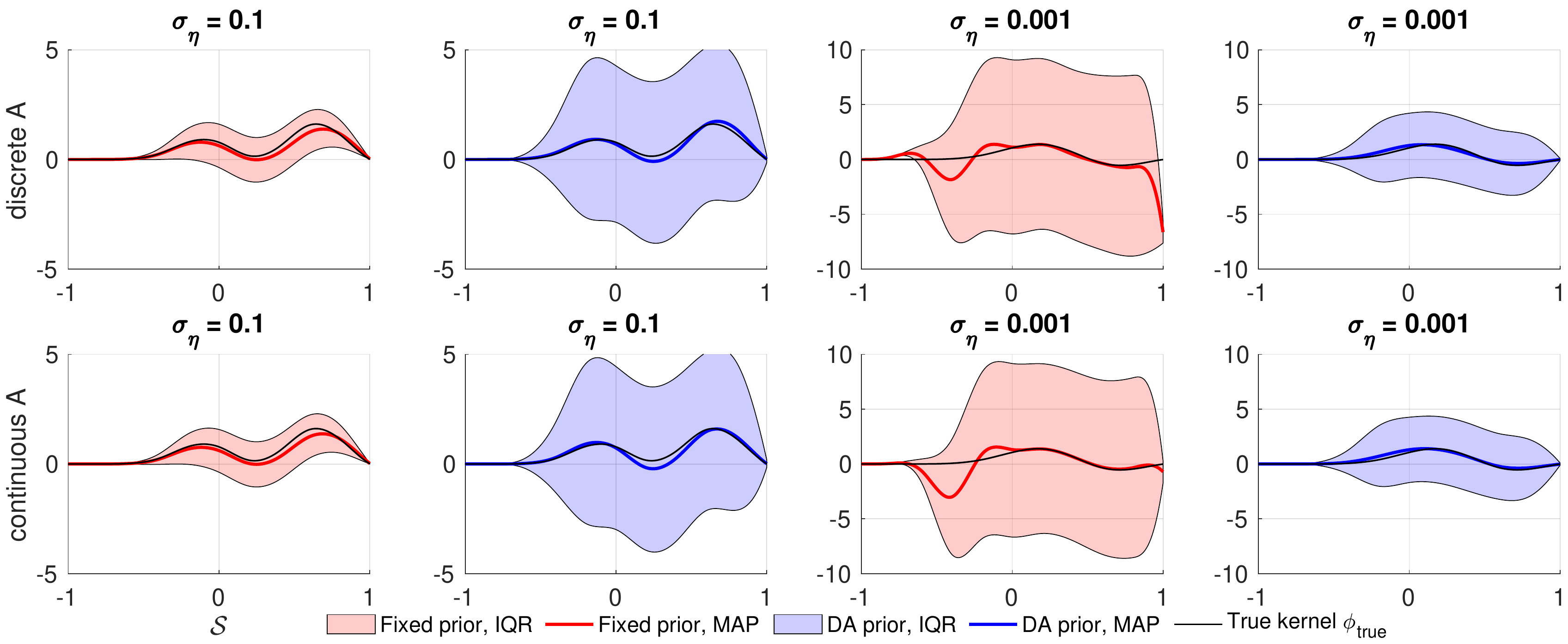}
\label{fig:post_in_FSOI}\vspace{-3mm} 
\ifjournal \vspace{-3mm} \fi 
\end{figure}

Figure \ref{fig:int_opt} shows the IQR of these simulations in the scenario that the true kernels are outside the FSOI. The fixed prior leads to diverging posterior means in 6 out of the 8 cases, while the DA-prior has stable posterior means in all cases. The fixed prior leads to a diverging posterior mean when using the continuously integrated regression matrix $\Abar_C$, because the discrepancy between $\bbar$ and $\Abar_C$ leads to a perturbation outside the FSOI, satisfying Assumption \ref{assumption2} (A3). Similarly, either the model error or partial observation error in $\bbar$ causes a perturbation outside the FSOI of $\Abar_D$, making the fixed prior's posterior mean diverge. On the other hand, the discretely computed $\Abar_D$ matches $\bbar$ in the sense that $\bbar\in {\rm Range}(\Abar_D)$ as proved in Proposition \ref{prop:lse-in-FSOI}, so the fixed prior has a stable posterior mean in cases of discretization and wrong noise assumption. In all these cases, the error of the posterior mean of the DA-prior does not decay as $\sigma_\eta\to 0$, because the error is dominated by the part outside of the FSOI that cannot be recovered from data.
\\
Figure \ref{fig:int_opt_in_FSOI} shows the IQR of these simulations with the true kernels sampled inside the FSOI. The DA prior leads to posterior means that are not only stable but also converge to small noise limits, whereas the fixed prior leads to diverging posterior means as in Figure \ref{fig:int_opt}. The convergence of the posterior means of the DA prior can be seen in the cases of ``Discretization'' and ``Wrong noise'' with both continuously and discretely computed regression matrices. Meanwhile, the flat lines of the DA prior in the cases of ``Model error'' or ``Partial observations'' are due to the error inside the FSOI caused by either the model error or partial observation error in $\bbar$, as shown in the proof of Theorem \ref{thm:MAP_stability}. 
\\
 Additionally, we show in Figure \ref{fig:post_outside_FSOI} and Figure \ref{fig:post_in_FSOI} the estimated posterior (in terms of its mean, the $75^{th}$ and $25^{th}$ percentiles) in a typical simulation, when $\phi_{true}$ is outside and inside the FSOI, respectively. Here, the percentiles are computed by drawing samples from the posterior. A more accurate posterior would have a more accurate mean and a narrower shaded region between the percentiles so as to have a smaller uncertainty. 
 In all cases, the DA prior leads to a more accurate posterior mean (MAP) than the fixed prior. 
 When the observation noise has $\sigma_\eta =0.1$, the DA prior leads to a posterior with a larger shaded region between the percentiles than the fixed prior, but when $\sigma_\eta= 0.001$, the DA prior's shaded region is much smaller than those of the fixed prior.   
 
In summary, these numerical results confirm that the data-adaptive prior removes the risk in a fixed non-degenerate prior,  leading to a robust posterior with a small noise limit.  
 \begin{remark}[Increasing data size] 
We have chosen a low number of data points in the numerical tests, so that the inverse problems are under-determined and hence are meaningful for a Bayesian study. In general, the blow-up behavior caused by a fixed non-degenerate prior will remain when the data size increases as long as the inverse problem satisfies the conditions in Assumption {\rm\ref{assumption2}}  (i.e., under-determined with deficient rank, using a fixed non-degenerate prior, with error in the null space of the normal operator). 
 \end{remark}

\subsection{Limitations of the data-adaptive prior}\label{sec:discussion}
As stated in \cite{hansen_LcurveIts_a}: ``\emph{every practical method has advantages and disadvantages}''.  The major advantage of the data-adaptive RKHS prior is to avoid the posterior being contaminated by the errors outside of the data-dependent function space of identifiability. 

A drawback of the data-adaptive prior is its reliance on selecting the hyper-parameter $\lambda_*$. The L-curve method is state-of-the-art and works well in our numerical tests, yet it has limitations in dealing with smoothness and asymptotic consistency (see \cite{hansen_LcurveIts_a}).  An improper hyper-parameter can lead to a posterior with an inaccurate mean and unreliable covariance. Also, the premise is that the identifiable part of the true kernel is in the data-adaptive RKHS. But this RKHS can be restrictive when the data is smooth, leading to an overly-smoothed estimator if the true kernel is non-smooth. It remains open to use a prior with $\LGbar^{s}$ as its covariance, as introduced in \cite{Lang2023small}, with $s\geq 0$ to detect the smoothness of the true kernel. We leave this as potential future work.

Also, the data-adaptive prior in this study is for linear inverse problems, and it does not apply to nonlinear problems in which the operator depends on the kernel nonlinearly. However, the covariance of our data-adaptive prior corresponds to a scaled Fisher information matrix. Thus, for nonlinear inverse problems, a potential adaptive prior is the scaled Fisher information, which has been explored as a regularization method in \cite{li2020_FisherInformation}.

\section{Conclusion} \label{sec:conc}
The inverse problem of learning kernels in operators is often severely ill-posed. We show that a fixed non-degenerate prior leads to a divergent posterior mean when the observation noise becomes small if the data induces a perturbation in the eigenspace of zero eigenvalues of the normal operator. 

We have solved the issue by a data-adaptive RKHS prior. It leads to a stable posterior whose mean always has a small noise limit, and the small noise limit converges to the identifiable part of the true kernel when the perturbation vanishes. Its covariance is the normal operator with a hyper-parameter selected adaptive to data by the L-curve method. Also, the data-adaptive prior improves the quality of the posterior over the fixed prior in two aspects: a smaller expected mean square error of the posterior mean, and a smaller trace of the covariance operator, thus reducing the uncertainty. 

Furthermore, we provide a detailed analysis of the data-adaptive prior in computational practice. We demonstrate its advantage on Toeplitz matrices and integral operators in the presence of four types of errors. Numerical tests show that when the noise becomes small, a fixed non-degenerate prior may lead to a divergent posterior mean, the data-adaptive prior always attains stable posterior means. 

We have also discussed the limitations of the data-adaptive prior,  such as its dependence on the hyper-parameter selection and its tendency to over-smoothing. It is of interest to overcome these limitations in future research by adaptively selecting the regularity of the prior covariance through a fractional operator. Among various other directions to be further explored, we mention one that is particularly relevant in the era of big data: to investigate the inverse problem when the data $\{u^k\}$ are randomly sampled in the setting of infinite-dimensional statistical models (e.g., \cite{gine2015mathematical}). When the operator $R_\phi[u]$ is linear in $u$, the examples of Toeplitz matrices and integral operators show that the inverse problem will become less ill-posed when the number of linearly independent data $\{u^k\}$ increases. When $R_\phi$ is nonlinear in $u$, it remains open to understand how the ill-posedness depends on the data. Another direction would be to consider sampling the posterior exploiting MCMC or sequential Monte Carlo methodologies (e.g., \cite{Robert}).

\section*{Acknowledgments}
The work of FL is funded by the Johns Hopkins University Catalyst Award, FA9550-20-1-0288, and NSF-DMS-2238486. XW is supported by the Johns Hopkins University research fund. FL would like to thank Yue Yu and Mauro Maggioni for inspiring discussions. 

\appendix
\section{Appendix}

\subsection{Identifiability theory}\label{append_ID}

The main theme in the identifiability theory is to find the function space in which the quadratic loss functional has a unique minimizer. 
The next lemma shows that the normal operator $\LGbar$ defined in \eqref{eq:LG} is a trace-class operator. Recall that an operator $\calQ$ on a Hilbert space if it satisfies $\sum_k\innerp{\calQ e_k,e_k}< \infty$ for any complete orthonormal basis $\{e_k\}_{k=1}^\infty$. 
\begin{lemma}\label{lemma:trace-opt}
Under Assumption {\rm \ref{assumption1}}, the operator $\LGbar:L^2_\rho\to L^2_\rho$ defined in \eqref{eq:LG} is a trace-class operator with $Tr(\LGbar) = \int_\calS  \Gbar(r,r) \rho(r)dr$.   
\end{lemma}
\begin{proof}
We have $\rho(r) = \frac{1}{ZN}\sum_{1\leq k\leq N}\int_\Omega  \left| g[u^k](x,r+x) \right|\mu(dx)$ by \eqref{eq:rho_conti}. Then,
\begin{equation*}
G(r,s)= \frac{1}{N}\sum_{1\leq k\leq N}\int  g[u^k](x,r+x)  g[u^k](x,s+x) \mu(dx) \leq  C \rho(r) \wedge \rho(s)
\end{equation*} 
for and $r,s\in \calS$, where $C= Z \max_{1\leq k\leq K}\sup_{x,y\in \Omega} |g[u^k](x,y)|$. Thus, $$\Gbar(r,s) = \frac{G(r,s)}{\rho(r)\rho(s)} \leq C\rho(r)^{-1}  \wedge \rho(s)^{-1},$$ for each $r,s\in \calS$. Meanwhile, since $\Omega$ is bonded, we have $|\calS|<\infty$. Hence $\int_\calS  \Gbar(r,r) \rho(r)dr \leq C |\calS| <\infty $.  Also, note that $\Gbar$ is continuous since $g[u^k]$ is continuous. Then, by \cite[Theorem 12, p344]{Lax02}, the operator $\LGbar$ with integral kernel $\Gbar$ has a finite trace $Tr(\LGbar) = \int_\calS  \Gbar(r,r) \rho(r)dr<\infty$. 
\end{proof}


 Theorem {\rm \ref{thm:FSOI}} characterizes the FSOI through the normal operator $\LGbar$.

\begin{proof}[Proof of Theorem {\rm \ref{thm:FSOI}}] 
Part $(a)$ follows from the definition of $\phi^\data$ in \eqref{eq:phi_f_N}. In fact, plugging in $f^k=R_{\phi_{true}}[u^k] + \xi_k + \eta_k$ into the right hand side of \eqref{eq:phi_f_N}, we have, $\forall \psi\in L^2_\rho$, 
\begin{align*}
\innerp{\phi^\data,\psi}_{L^2_\rho} 
& = \frac{1}{N}\sum_{1\leq k\leq N}\innerp{ R_\psi[u^k], R_{\phi_{true}}[u^k] }_{\spaceY} + \innerp{ R_\psi[u^k], \xi_k] }_{\spaceY} + \innerp{ R_\psi[u^k], \eta_k] }_{\spaceY} \\
& = \innerp{\psi,\LGbar\phi_{true}}_{L^2_\rho} + \innerp{\psi,\epsilon^\xi}_{L^2_\rho} +  \innerp{\psi,\epsilon^\eta}_{L^2_\rho},
\end{align*}
where the first term in the last equation comes from the definitions of the operator $\LGbar$ in \eqref{eq:LG}, the second and the third term comes from the Riesz representation. Since each $\eta_k$ is a $\spaceY$-valued white noise, the random variable $ \innerp{\psi,\epsilon^\eta}_{L^2_\rho} =  \frac{1}{N}\sum_{1\leq k\leq N}\innerp{ R_\psi[u^k], \eta_k] }_{\spaceY} $ is Gaussian with mean zero and variance $\sigma_\eta^2\innerp{\psi,\LGbar\psi}_{L^2_\rho}$ for each $\psi \in L^2_\rho$. Thus, $\epsilon^\eta $ has a Gaussian distribution $\calN(0, \sigma_\eta^2\LGbar )$.

Part $(b)$ follows directly from loss functional in \eqref{eq:lossFn2}. 

For Part $(c)$, first, note that the quadratic loss functional has a unique minimizer in $H$. Meanwhile, note that $H$ is the orthogonal complement of the null space of $\LGbar$, and $\calE(\phi_{true}+\phi^0) = \calE(\phi_{true})$ for any $\phi^0$ such that $\LGbar \phi^0=0$.  Thus, $H$ is the largest such function space, and we conclude that  $H$ is the FSOI.
 
Next, for any $\phi^\data\in \LGbar(L^2_\rho)$, the estimator $\widehat{\phi} = \LGbar^{-1}\phi^\data$ is well-defined. By Part $(b)$, this estimator is the unique zero of the loss functional's Fr\'echet derivative in $H$. Hence it is the unique minimizer of $\calE(\phi)$ in $H$. In particular, when the data is noiseless and with no model error, and it is generated from $\phi_{true}$, i.e. $R_{\phi_{true}}[u^k]=f^k$, we have $\phi^\data= \LGbar\phi_{true}$ from Part $(a)$. Hence $\widehat{\phi} = \LGbar^{-1}\phi^\data = \phi_{true}$. That is, $\phi_{true}\in H$ is the unique minimizer of the loss functional $\calE$. 
\end{proof}

The proof of Proposition {\rm\ref{lemma:LG-A}} is an extension of Theorem 4.1 of {\rm\cite{LLA22}}. 
\begin{proof}[Proof of Proposition {\rm\ref{lemma:LG-A}}] Let $\psi_k = \sum_{j=1}^l  V_{jk}\phi_j$ with $V^\top BV  = I$.  Then, $\psi_k$ is an eigenfunction of $\LGbar$ with eigenvalue $\lambda_k$ if and only if for each $i$, 
\begin{align*}
 \innerp{\phi_i, \lambda_k \psi_k}_{L^2_\rho} = \innerp{\phi_i, \LGbar \psi_k}_{L^2_\rho} =\sum_{1\leq j\leq l}  \innerp{\phi_i,\LGbar \phi_j}_{L^2_\rho}  V_{jk}= \sum_{1\leq j\leq l} \Abar (i,j) V_{jk},\vspace{-2mm} 
\end{align*}
where the last equality follows from the definition of $\Abar$. 
Meanwhile, by the definition of $B$ we have $ \innerp{\phi_i, \lambda_k \psi_k}_{L^2_\rho} = \sum_{j=1}^l B(i,j)  V_{jk}\lambda_k$ for each $i$. Then, Equation \eqref{eq:gEigenP} follows. 

Next, to compute $\innerp{\phi,\LGbar^{-1}\phi}_{L^2_\rho} $, we denote $\Psi = (\psi_1,\ldots,\psi_l)^\top$ and $\Phi = (\phi_1,\ldots,\phi_l)^\top$. Then, we can write  
$$ \Psi= V^\top \Phi,  \qquad \phi = \sum_{1\leq i \leq l} c_i\phi_i = c^\top \Phi =  c^\top V^{-\top} \Psi.$$
Hence, we can obtain $B_{rkhs}= (V\Lambda V^\top)^{-1}$ in $\innerp{\phi,\LGbar^{-1}\phi}_{L^2_\rho} = c^\top B_{rkhs} c $ via: 
\begin{align*}
\innerp{\phi,\LGbar^{-1}\phi}_{L^2_\rho}  &= \innerp{c^\top \Phi,\LGbar^{-1}c^\top\Phi }_{L^2_\rho} \notag
\\&=  \innerp{c^\top V^{-\top} \Psi,\LGbar^{-1}c^\top V^{-\top} \Psi }_{L^2_\rho} \notag  \\
& =c^\top V^{-\top} \innerp{ \Psi,\LGbar^{-1} \Psi }_{L^2_\rho}  V^{-1} c =   c^\top V^{-\top} \Lambda^{-1} V^{-1} c,   \notag
\end{align*}
where the last equality follows from $\innerp{ \Psi,\LGbar^{-1} \Psi }_{L^2_\rho} = \Lambda^{-1}$.

Additionally, to prove $B_{rkhs} = B\Abar^{-1}B$, we use the generalized eigenvalue problem. Since  $V^\top B V = I$, we have $V^{-1} =V^\top B$.  Meanwhile,  $\Abar V= B  V\Lambda$  implied that $B^{-1}\Abar = V\Lambda V^{-1}$. Thus, $B^{-1}\Abar B^{-1} = V\Lambda V^{-1} = V\Lambda V^\top  $, which is $B_{rkhs}^{-1}$. 
\end{proof}

\subsection{Gaussian measures on a Hilbert space} \label{sec:gaussian_Hilbert}
A Gaussian measure on a Hilbert space is defined by its mean and covariance operator (see \cite[Chapter 1-2]{da2006introduction} and \cite{da2014stochastic}). Let  $H$ be a Hilbert space with inner product $\innerp{\cdot,\cdot}$, and let $\mathcal{B}(H)$ denote its Borel algebra. Let $\calQ$ be a symmetric nonnegative trace class operator on $H$, that is $\innerp{\calQ x,y}  = \innerp{x,\calQ y}$ and $\innerp{\calQ x,x}\geq 0$ for any $x,y\in H$, and $\sum_k\innerp{\calQ e_k,e_k}< \infty$ for any complete orthonormal basis $\{e_k\}_{k=1}^\infty$. Additionally, denote $\{\lambda_k,e_k\}_{k=1}^\infty$ the eigenvalues (in descending order) and eigenfunctions of $\calQ$.

A measure on $H$ with mean $a$ and covariance operator $\calQ$ is a Gaussian measure $\pi = \mathcal{N}(a,\calQ)$ iff its Fourier transform $\widehat \pi(h)= \int_H e^{i\innerp{x,h}} \pi(dx)$ is $e^{i\innerp{a,h} - \frac{1}{2}\innerp{\calQ h,h}}$ for any $h\in H$. 
The measure is non-degenerate if $\mathrm{Ker} \calQ = \{0\}$, i.e., $\lambda_k>0 $ for all $k$. It is a product measure $\pi = \prod_{k=1}^\infty \calN(a_k,\lambda_k)$, where $a_k= \innerp{a,e_k} \in \R$ for each $k$.  Note that $\pi(\calQ^{1/2}H) = 0$ if $H$ is infinite-dimensional, that is, the Cameron-Martin space $\calQ^{1/2}H$ has measure zero. 

The next lemma specifies the covariance of the coefficient of an $H$-valued Gaussian random variable. The coefficient can be on either a full or partial basis. 
\begin{lemma}[Covariance operator to covariance matrix]\label{lemma:Gausian_basis} 
Let $H$ be a Hilbert space with a complete basis $\{\phi_i\}_{i=1}^n$ that may not be orthonormal, where $n\leq \infty$. For $l\leq n$, assume that the matrix $B=\innerp{\phi_i,\phi_j}_{1\leq i,j\leq l}$ is strictly positive definite. Let $\phi=\sum_{i=1}^n c_i\phi_i $ be an $H$-valued random variable with Gaussion measure $\calN(m,\calQ)$, where $\calQ$ is a trace class operator. Then, the coefficient $c = (c_1,\ldots, c_l)^\top\in \R^{l\times 1}$ has a Gaussian distribution $\calN(\overline c,B^{-1} A B^{-1})$, where 
\begin{align*}
\overline c & = B^{-1} (\innerp{\phi,\phi_1}, \ldots,\innerp{\phi,\phi_l})^\top \\
A               & = [A(i,j)]_{1\leq i,j\leq l} = [\innerp{\phi_i,\calQ \phi_j}]_{1\leq i,j\leq l}.
\end{align*}
\end{lemma}
\begin{proof} Without loss of generality, we assume $m=0$: otherwise, we replace $\phi$ by $\phi-m$ in the following proof. By definition, for any $h\in \mH$, the random variable $\innerp{\phi,h}$ has distribution $\calN(0,\innerp{h,\calQ h})$. Thus, we have $\innerp{\phi,\phi_i} \sim \calN(0,\innerp{\phi_i,\calQ \phi_i})$ for each $i$. Similarly, we have that  $\E[\innerp{\phi,\phi_i+\phi_j}^2] =\innerp{\phi_i+\phi_j,\calQ (\phi_i+\phi_j)} $. Then, we have 
\begin{align*}
	\E[\innerp{\phi,\phi_i}\innerp{\phi,\phi_j}] &=  \frac{1}{2}\Big(\E[\innerp{\phi,\phi_i+\phi_j}^2] - \E[\innerp{\phi,\phi_i}^2] - \E[\innerp{\phi,\phi_j}^2]\Big) \\
	&=\frac{1}{2}\Big( \innerp{\phi_i+\phi_j,\calQ (\phi_i+\phi_j)}-\innerp{\phi_i,\calQ \phi_i}-\innerp{\phi_j,\calQ \phi_j}\Big)= \innerp{\phi_i,\calQ \phi_j} .
\end{align*}
  Hence, the random vector $X= (\innerp{\phi,\phi_1},\ldots, \innerp{\phi,\phi_l} )^\top $ is  Gaussian $\calN(0, A)$ with $A(i,j) = \innerp{\phi_i,\calQ \phi_j}$. Now, noticing that $X=  Bc$ and $B= B^\top$, we obtain that the distribution of $c= B^{-1}X$ is $\calN(0,B^{-1} A B^{-1})$, where the covariance matrix can be derived as $\E[cc^\top] = \E[B^{-1}XX^\top B^{-1}] = B^{-1}AB^{-1}$.
\end{proof}

On the other hand, the distribution of the coefficient only determines a Gaussian measure on the linear space its basis spans. 
\begin{lemma}[Covariance matrix to covariance operator]\label{lemma:basis_to_operator}
 Let $H = \mathrm{span}\{\phi_i\}_{i=1}^l$ with $l\leq \infty$ be a Hilbert space with basis such that the matrix $B=\innerp{\phi_i,\phi_j}_{1\leq i,j\leq l}$ is strictly positive definite.  Assume the coefficient $c\in \R^l$ of $\phi=\sum_{i=1}^l c_i\phi_i $ has a Gaussian measure $\calN(0,Q)$. Then, the $H$-valued random variable $\phi$ has a Gaussian distribution $\calN(0,\calQ)$, where the operator $\calQ$ is defined by $\innerp{\phi_i,\calQ \phi_j} = (BQB)_{i,j}$.  
\end{lemma}
\begin{proof} Since $\{\phi_i\}$ is a complete basis, we only need to determine the distribution of the random vector $X= (\innerp{\phi,\phi_1},\ldots, \innerp{\phi,\phi_l} )^\top \in \R^l$. Note that it satisfies $X=  Bc$. Thus, its distribution is Gaussian $\calN(0,BQB)$. 
\end{proof}

\subsection{Details of numerical examples}\label{appd-computeToeplitz}
\paragraph{Computation for Toeplitz matrix.} 
Each dataset $\{u^{k}=(u_0^k,u_1^k)\}_k$ leads to an exploration measure  on  $\calS=\{-1,0,1\}$: 
\begin{equation}\label{eq:rho1vector} \notag
\rho(-1) = \frac{\sum_k |u_1^k|}{2\sum_k (|u_1^k|+|u_0^k|)}, \quad \rho(0) = \frac{1}{2}, \quad \rho(1) =\frac{\sum_k |u_0^k|}{2\sum_k (|u_1^k|+|u_0^k|)} .
\end{equation} 
Since each $u=(u_0,u_1)$ leads to a rank-2 regression matrix
\begin{equation}\label{eq:Lu2d}\notag
L_u = \begin{bmatrix}
    u_1 & u_0 & 0 \\
    0 &  u_1 & u_0 
        \end{bmatrix}, \quad 
L_u^\top L_u= \begin{bmatrix}
    u_1^2 & u_1u_0 & 0 \\
    u_1u_0 &  u_1^2+u_0^2 & u_1u_0 \\
    0 & u_1u_0 & u_0^2 
        \end{bmatrix}, 
\end{equation}
the regression matrices $\Abar= \sum_{k} L_{u^k}^\top L_{u^k}$ of the three datasets are 
\begin{equation}\label{eq:A_toeplitz}
\Abar_{(1)}= \begin{bmatrix}     0 & 0 & 0 \\     0 &  1 & 0\\     0 & 0 & 1          \end{bmatrix}, \quad  
  \Abar_{(2)} = \frac{1}{2}\sum_{k=1}^2 L_{u^k}^\top L_{u^k}=\frac 12 \begin{bmatrix}     1 & 0 & 0 \\     0 &  2 & 0\\     0 & 0 & 1           \end{bmatrix}, \quad
  \Abar_{(3)}= \begin{bmatrix}     1 & 1& 0 \\     1 &  2 & 1\\     0 & 1 & 1         \end{bmatrix}. 
\end{equation}
Additionally, with $B= \mathrm{Diag}(\rho)$, the prior covariances $\lambda_* Q_0^\data = B^{-1} \Abar B^{-1}$ are 
\begin{equation}\label{eq:Qdata_toeplitz}
Q^\data_{0,(1)}= \begin{bmatrix}     0 & 0 & 0 \\     0 &  4 & 0\\     0 & 0 & 4          \end{bmatrix}, \quad  
Q^\data_{0,(2)}
= \begin{bmatrix}     8 & 0 & 0 \\     0 &  4 & 0\\     0 & 0 & 8           \end{bmatrix}, \quad 
Q^\data_{0,(3)}= \begin{bmatrix}    16 & 8 & 0 \\     8 &  8 & 8\\     0 & 8 & 16        \end{bmatrix}.
\end{equation}

We analyze the well-posedness of the inverse problem in terms of the operator $\LGbar$, whose eigenvalues are solved via the generalized eigenvalue problem (see  Proposition \ref{lemma:LG-A}). 
\begin{itemize}
\item 	For the data set $\{u^1\}$, the exploration measure $\rho$ is degenerate with $\rho(-1)=0$, thus, we have no information from data to identify $\phi(-1)$. As a result, $L^2_\rho=\mathrm{span}\{\phi_2,\phi_3\} $ is a proper subspace of $\R^3$. The regression matrix $\Abar_{(1)}$ and the covariance matrix $Q^\data_{0,(1)}$ are effectively the identity matrix $I_2$ and $4I_2$. The operator $\LGbar$ has eigenvalues $\{1,1\}$, and the FSOI is $L^2_\rho$. Thus, the inverse problem is well-posed in $L^2_\rho$. 
\item 
For the dataset $\{u^1,u^2\}$, the inverse problem is well-posed because the operator $\LGbar$ has eigenvalues $\{2,2,2\}$, and the FSOI is $L^2_\rho$. Note that the data-adaptive prior $Q^\data_{0,(2)}$ assigns weights to the entries of the coefficient according to the exploration measure. 
\item 
For the data set $\{u^3\}$, the inverse problem is \emph{ill-defined in $L^2_\rho$}, but it is \emph{well-posed in the FSOI}, which is a proper subset of $L^2_\rho$. Here the FSOI is $\mathrm{span}\{\psi_1, \psi_2\}$, which are the eigenvectors of $\LGbar$ with positive eigenvalues. Following \eqref{eq:gEigenP}, these eigenvectors $\{\psi_i\}$ are solved from the generalized eigenvalue problem $\Abar_{(3)} \psi =\lambda \mathrm{Diag}(\rho) \psi $ and they are orthonormal in $L^2_\rho$. The eigenvalues are $\{8,4,0\}$ and the corresponding eigenvectors are $\psi_1 = (1,1,1)^\top$, $\psi_2 = (-\sqrt{2},0,\sqrt{2})^\top,$ and $ \psi_3 = (1,-1,1)^\top$. 

\end{itemize}

\begin{figure}[thb!]
\caption{The L-curve for selecting the hyper-parameter $\lambda_*$. 
} \label{fig:Lcurve}
\vspace{-2mm}
\centering
 \includegraphics[width=0.6\textwidth]{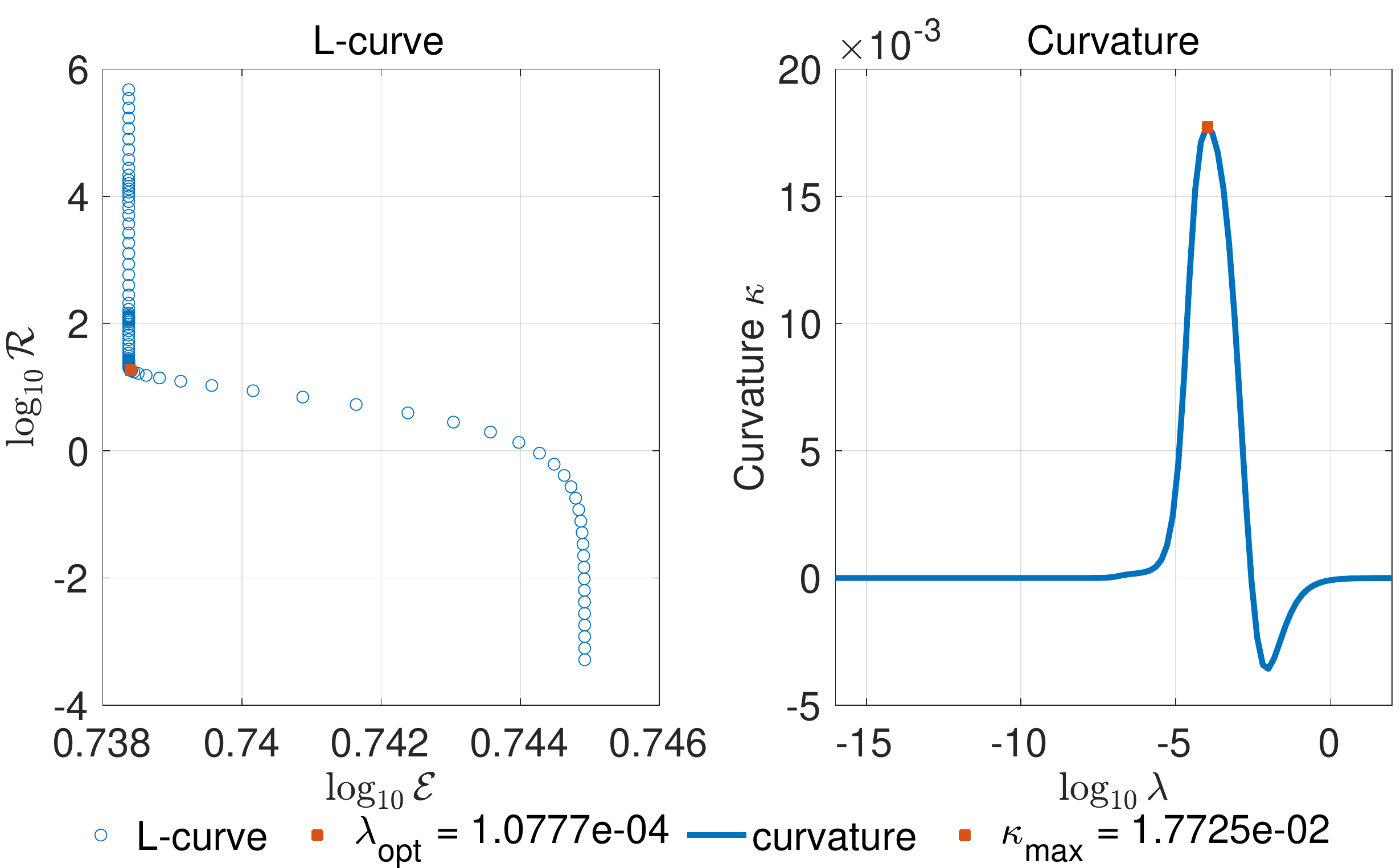}
 \vspace{-2mm}
\end{figure}

\paragraph{The hyper-parameter selected by the L-curve method.}
 Figure \ref{fig:Lcurve} shows a typical L-curve, where $\mathcal{R}(\lambda)=\norm{\phi_\lambda}_{H_G}$ and $\mathcal{E}$ represents the square root of the loss $\mathcal{E}(\phi_\lambda)$. The L-curve method selects the parameter that attains the maximal curvature at the corner of the L-shaped curve.

\begin{figure}[thb!]
\caption{The hyper-parameter $\lambda_*$ in the 200 simulations in Figure \ref{fig:int_opt}. 
} \label{fig:lambda_stats}
\vspace{-2mm}
\centering
 \includegraphics[width=0.98\textwidth]{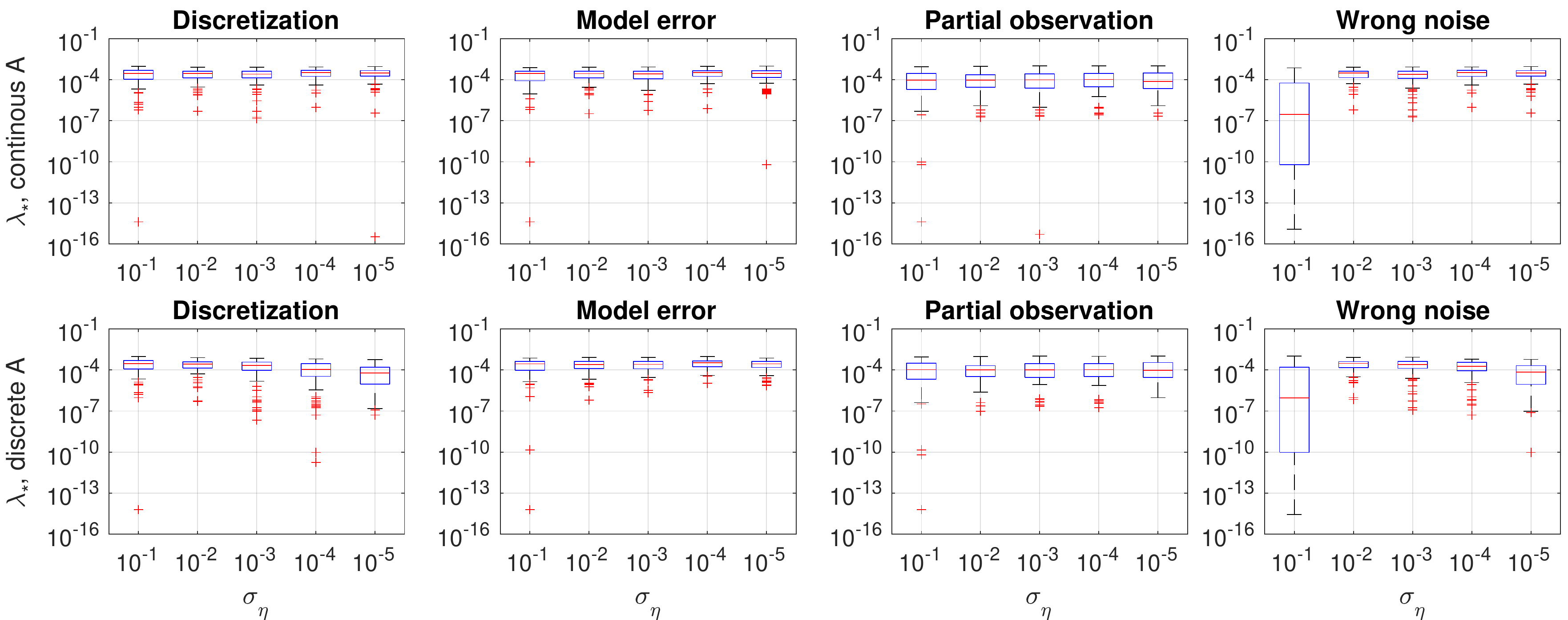}
 \vspace{-3mm}
\end{figure}

\begin{figure}[thb!]
\caption{The hyper-parameter $\lambda_*$ in the 200 simulations in Figure \ref{fig:int_opt_in_FSOI}. 
} \label{fig:lambda_stats_inFSOI}
\vspace{-2mm}
\centering
\includegraphics[width=0.98\textwidth]{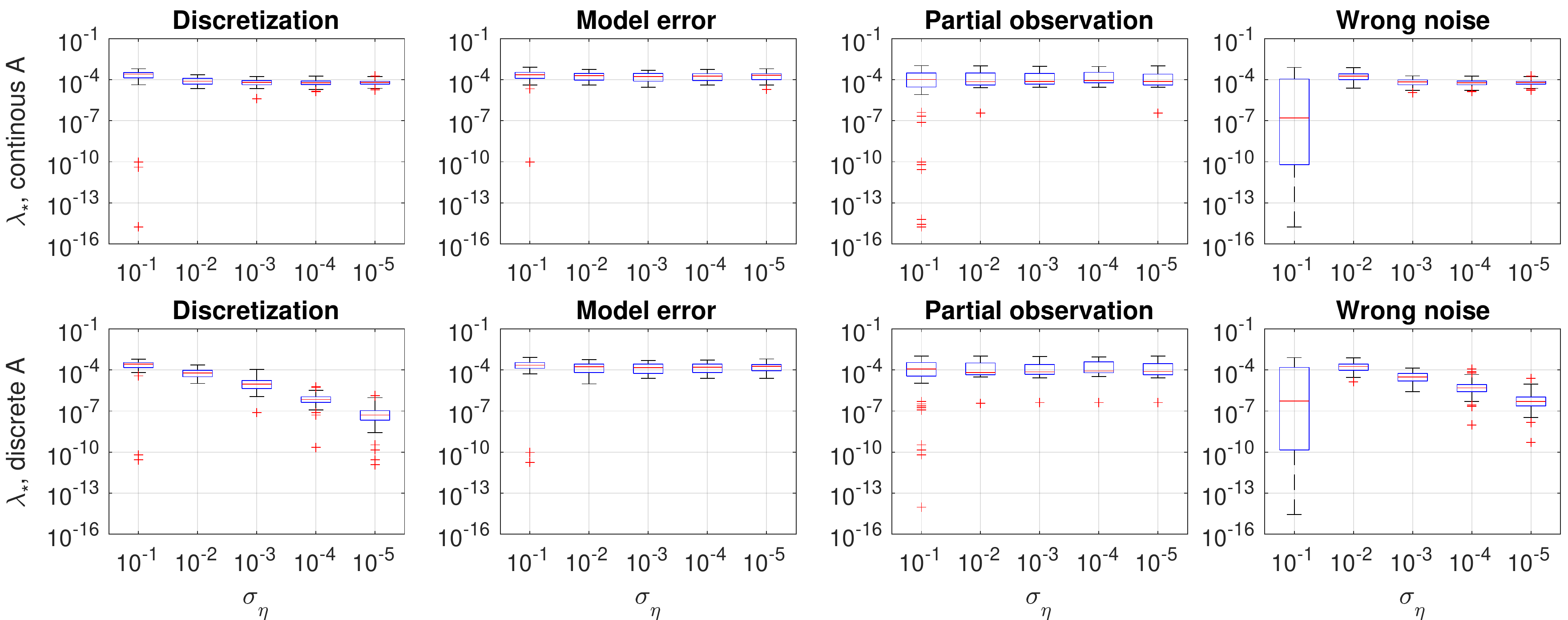}
 \vspace{-3mm}
\end{figure}

Figures \ref{fig:lambda_stats}--\ref{fig:lambda_stats_inFSOI} present the $\lambda_*$ in the simulations in Figures  \ref{fig:int_opt}--\ref{fig:int_opt_in_FSOI}, respectively. Those hyper-parameters are mostly similar, and the majority of them are at the scale of $10^{-4}$. They show that the optimal hyper-parameter depends on the spectrum of $\LGbar$, the four types of errors in $\bbar$, the strength of the noise, and the smoothness of the true kernel. 
Generally, a large variation of $\lambda_*$ suggests difficulty in selecting an optimal hyper-parameter by the method. 
Additionally, the error in the numerical computation of matrix inversion or the solution of the linear systems can affect the result when $\lambda_*$ is small. Thus, it is beyond the scope of this study to analyze the optimal hyper-parameter. 

\bigskip  \smallskip

\ifarXiv 
\bibliographystyle{myplain}
\fi
{\small 
\bibliography{ref_regularization,ref_FeiLU2022_10,ref_prior_Bayesian,ref_OED,ref_nonlocal_kernel,ref_sparseRegression,ref_kernel_methods,ref_IPS_learning}

\begin{thebibliography}{10}

\bibitem{Agliari1988gprior}
A. Agliari and C.~C. Parisetti.
\newblock A-g reference informative prior: A note on zellner's $g$ prior.
\newblock {\em Journal of the Royal Statistical Society. Series D (The
  Statistician)}, 37(3):271--275, 1988.

\bibitem{alexanderian2016bayesian}
A. Alexanderian, P.~J. Gloor, and O. Ghattas.
\newblock On bayesian a-and d-optimal experimental designs in infinite
  dimensions.
\newblock {\em Bayesian Analysis}, 11(3):671--695, 2016.

\bibitem{bauer2007regularization}
F. Bauer, S. Pereverzev, and L. Rosasco.
\newblock On regularization algorithms in learning theory.
\newblock {\em Journal of complexity}, 23(1):52--72, 2007.

\bibitem{Bayarri2012Criteria}
M.~J. Bayarri, J.~O. Berger, A. Forte, and G. Garc{\'\i}a-Donato.
\newblock Criteria for bayesian model choice with application to variable
  selection.
\newblock {\em The Annals of Statistics}, 40(3), Jun 2012.

\bibitem{belkin2018understand}
M. Belkin, S. Ma, and S. Mandal.
\newblock To understand deep learning we need to understand kernel learning.
\newblock In {\em International Conference on Machine Learning}, pages
  541--549. PMLR, 2018.

\bibitem{bucur2016_NonlocalDiffusion}
C. Bucur and E. Valdinoci.
\newblock {\em Nonlocal {{Diffusion}} and {{Applications}}}, volume~20 of {\em
  Lecture {{Notes}} of the {{Unione Matematica Italiana}}}.
\newblock {Springer International Publishing}, {Cham}, 2016.

\bibitem{carrillo2019aggregation}
J.~A. Carrillo, K. Craig, and Y. Yao.
\newblock Aggregation-diffusion equations: dynamics, asymptotics, and singular
  limits.
\newblock In {\em Active Particles, Volume 2}, pages 65--108. Springer, 2019.

\bibitem{chaloner1995bayesian}
K. Chaloner and I. Verdinelli.
\newblock Bayesian experimental design: A review.
\newblock {\em Statistical Science}, pages 273--304, 1995.

\bibitem{CZ07book}
F. Cucker and D.~X. Zhou.
\newblock {\em Learning theory: an approximation theory viewpoint}, volume~24.
\newblock Cambridge University Press, {Cambridge}, 2007.

\bibitem{cui2022unified}
T. Cui and X.~T. Tong.
\newblock A unified performance analysis of likelihood-informed subspace
  methods.
\newblock {\em Bernoulli}, 28(4):2788--2815, 2022.

\bibitem{da2006introduction}
G. Da~Prato.
\newblock {\em An introduction to infinite-dimensional analysis}.
\newblock Springer Science \& Business Media, {New York}, 2006.

\bibitem{da2014stochastic}
G. Da~Prato and J. Zabczyk.
\newblock {\em Stochastic equations in infinite dimensions}.
\newblock Cambridge university press, 2014.

\bibitem{darcy2021learning}
M. Darcy, B. Hamzi, J. Susiluoto, A. Braverman, and H. Owhadi.
\newblock Learning dynamical systems from data: a simple cross-validation
  perspective, part ii: nonparametric kernel flows.
\newblock {\em preprint}, 2021.

\bibitem{dashti2017bayesian}
M. Dashti and A.~M. Stuart.
\newblock The {Bayesian} approach to inverse problems.
\newblock In {\em Handbook of uncertainty quantification}, pages 311--428.
  Springer, 2017.

\bibitem{HHQ22}
M.~V. de~Hoop, D.~Z. Huang, E. Quin, and A.~M. Stuart.
\newblock The cost-accuracy trade-off in operator learning with neural
  networks.
\newblock {\em arXiv preprint arXiv:2203.13181}, 2022.

\bibitem{HKN22}
M.~V. de~Hoop, N.~B. Kovachki, N.~H. Nelsen, and A.~M. Stuart.
\newblock Convergence rates for learning linear operators from noisy data.
\newblock {\em SIAM/ASA Journal on Uncertainty Quantification}, 2022.

\bibitem{delia2020_NumericalMethods}
M. D'Elia, Q. Du, C. Glusa, M. Gunzburger, X. Tian, and Z. Zhou.
\newblock Numerical methods for nonlocal and fractional models.
\newblock {\em Acta Numerica}, 29:1--124, 2020.

\bibitem{della2022nonparametric}
L. Della~Maestra and M. Hoffmann.
\newblock Nonparametric estimation for interacting particle systems:
  {{McKean}}-{{Vlasov}} models.
\newblock {\em Probability Theory and Related Fields}, pages 1--63, 2022.

\bibitem{du2012_AnalysisApproximation}
Q. Du, M. Gunzburger, R.~B. Lehoucq, and K. Zhou.
\newblock Analysis and {{Approximation}} of {{Nonlocal Diffusion Problems}}
  with {{Volume Constraints}}.
\newblock {\em SIAM Rev.}, 54(4):667--696, 2012.

\bibitem{FRT21-GP}
J. Feng, Y. Ren, and S. Tang.
\newblock Data-driven discovery of interacting particle systems using gaussian
  processes.
\newblock {\em arXiv preprint arXiv:2106.02735}, 2021.

\bibitem{gazzola2019ir}
S. Gazzola, P.~C. Hansen, and J.~G. Nagy.
\newblock Ir tools: a matlab package of iterative regularization methods and
  large-scale test problems.
\newblock {\em Numerical Algorithms}, 81(3):773--811, 2019.

\bibitem{gine2015mathematical}
E. Gin{\'e} and R. Nickl.
\newblock {\em Mathematical foundations of infinite-dimensional statistical
  models}, volume~40.
\newblock Cambridge University Press, {UK}, 2015.

\bibitem{hansen1994_regularization_tools}
P.~C. Hansen.
\newblock {{REGULARIZATION TOOLS}}: {{A Matlab}} package for analysis and
  solution of discrete ill-posed problems.
\newblock {\em Numer Algor}, 6(1):1--35, 1994.

\bibitem{hansen_LcurveIts_a}
P.~C. Hansen.
\newblock The {L}-curve and its use in the numerical treatment of inverse
  problems.
\newblock In {\em in Computational Inverse Problems in Electrocardiology, ed.
  P. Johnston, Advances in Computational Bioengineering}, pages 119--142. WIT
  Press, 2000.

\bibitem{he2022numerical}
Y. He, S.~H. Kang, W. Liao, H. Liu, and Y. Liu.
\newblock Numerical identification of nonlocal potential in aggregation.
\newblock {\em Commun. Comput. Phys.}, 32:638--670, 2022.

\bibitem{hofmann2008kernel}
T. Hofmann, B. Schölkopf, and A.~J. Smola.
\newblock Kernel methods in machine learning.
\newblock {\em Ann. Statist.}, 36(3):1171--1220, 06 2008.

\bibitem{jeffreys1998theory}
H. Jeffreys.
\newblock {\em The theory of probability}.
\newblock OuP Oxford, 1961.

\bibitem{KS05}
J. Kaipio and E. Somersalo.
\newblock {\em Statistical and Computational Inverse Problems}.
\newblock Springer, 2005.

\bibitem{KLA21}
N.~B. Kovachki, Z. Li, K. Azizzadenesheli, B. Liu, K. Bhattacharya, A.~M.
  Stuart, and A. Anandkumar.
\newblock Neural operator: Learning maps between function spaces.
\newblock {\em arXiv preprint arXiv:2108.08481}, 2021.

\bibitem{LangLu22}
Q. Lang and F. Lu.
\newblock Learning interaction kernels in mean-field equations of first-order
  systems of interacting particles.
\newblock {\em SIAM Journal on Scientific Computing}, 44(1):A260--A285, 2022.

\bibitem{Lang2023small}
Q. Lang and F. Lu.
\newblock Small noise analysis for {Tikhonov and RKHS} regularizations.
\newblock {\em arXiv preprint arXiv:2305.11055}, 2023.

\bibitem{Lax02}
P.~D. Lax.
\newblock {\em Functional Analysis}.
\newblock John Wiley \& Sons Inc., New York, 2002.

\bibitem{Lehtinen}
M.~S. Lehtinen, L. Paivarinta, and E. Somersalo.
\newblock Linear inverse problems for generalised random variables.
\newblock {\em Inverse Problems}, 5(4):599--612, 1989.

\bibitem{li2020_FisherInformation}
W. Li, J. Lu, and L. Wang.
\newblock Fisher information regularization schemes for {{Wasserstein}}
  gradient flows.
\newblock {\em Journal of Computational Physics}, 416:109449, 2020.

\bibitem{LLMTZ21}
Z. Li, F. Lu, M. Maggioni, S. Tang, and C. Zhang.
\newblock On the identifiability of interaction functions in systems of
  interacting particles.
\newblock {\em Stochastic Processes and their Applications}, 132:135--163,
  2021.

\bibitem{LKN20}
Z. Li, N.~B. Kovachki, K. Azizzadenesheli, B. Liu, K. Bhattacharya, A.~M.
  Stuart, and A. Anandkumar.
\newblock Fourier neural operator for parametric partial differential
  equations.
\newblock {\em International Conference on Learning Representations}, 2020.

\bibitem{LAY22}
F. Lu, Q. An, and Y. Yu.
\newblock Nonparametric learning of kernels in nonlocal operators.
\newblock {\em Journal of Peridynamics and Nonlocal Modeling}, pages 1--24,
  2023.

\bibitem{LLA22}
F. Lu, Q. Lang, and Q. An.
\newblock {Data adaptive RKHS Tikhonov regularization for learning kernels in
  operators}.
\newblock {\em Proceedings of Mathematical and Scientific Machine Learning,
  PMLR 190:158-172}, 2022.

\bibitem{LMT21_JMLR}
F. Lu, M. Maggioni, and S. Tang.
\newblock Learning interaction kernels in heterogeneous systems of agents from
  multiple trajectories.
\newblock {\em Journal of Machine Learning Research}, 22(32):1--67, 2021.

\bibitem{LMT21}
F. Lu, M. Maggioni, and S. Tang.
\newblock Learning interaction kernels in stochastic systems of interacting
  particles from multiple trajectories.
\newblock {\em Foundations of Computational Mathematics}, pages 1--55, 2021.

\bibitem{LZTM19pnas}
F. Lu, M. Zhong, S. Tang, and M. Maggioni.
\newblock Nonparametric inference of interaction laws in systems of agents from
  trajectory data.
\newblock {\em Proc. Natl. Acad. Sci. USA}, 116(29):14424--14433, 2019.

\bibitem{LJK19}
L. Lu, P. Jin, and G.~E. Karniadakis.
\newblock Deeponet: Learning nonlinear operators for identifying differential
  equations based on the universal approximation theorem of operators.
\newblock {\em arXiv preprint arXiv:1910.03193}, 2021.

\bibitem{LJP21}
L. Lu, P. Jin, G. Pang, Z. Zhang, and G.~E. Karniadakis.
\newblock Learning nonlinear operators via {deepOnet} based on the universal
  approximation theorem of operators.
\newblock {\em Nature Machine Intelligence}, 3(3):218--229, 2021.

\bibitem{mavridis2022learning}
C.~N. Mavridis, A. Tirumalai, and J.~S. Baras.
\newblock Learning swarm interaction dynamics from density evolution.
\newblock {\em IEEE Transactions on Control of Network Systems},
  10(1):214--225, 2022.

\bibitem{messenger2022learning}
D.~A. Messenger and D.~M. Bortz.
\newblock Learning mean-field equations from particle data using wsindy.
\newblock {\em Physica D: Nonlinear Phenomena}, 439:133406, 2022.

\bibitem{MT14}
S. Motsch and E. Tadmor.
\newblock {Heterophilious Dynamics Enhances Consensus}.
\newblock {\em SIAM Rev}, 56(4):577 -- 621, 2014.

\bibitem{owhadi2019kernel}
H. Owhadi and G.~R. Yoo.
\newblock Kernel flows: From learning kernels from data into the abyss.
\newblock {\em Journal of Computational Physics}, 389:22--47, 2019.

\bibitem{Robert}
C. {Robert} and G. Casella, editors.
\newblock {\em Monte Carlo Statistical Methods}.
\newblock {Springer}, 1999.

\bibitem{rudin1992nonlinear}
L.~I. Rudin, S. Osher, and E. Fatemi.
\newblock Nonlinear total variation based noise removal algorithms.
\newblock {\em Physica D: nonlinear phenomena}, 60(1-4):259--268, 1992.

\bibitem{shampine2008vectorized}
L.~F. Shampine.
\newblock Vectorized adaptive quadrature in matlab.
\newblock {\em Journal of Computational and Applied Mathematics},
  211(2):131--140, 2008.

\bibitem{spantini2015optimal}
A. Spantini, A. Solonen, T. Cui, J. Martin, L. Tenorio, and Y. Marzouk.
\newblock Optimal low-rank approximations of {Bayesian} linear inverse problems.
\newblock {\em SIAM Journal on Scientific Computing}, 37(6):A2451--A2487, 2015.

\bibitem{Sriperumbudur2011Universality}
B.~K. Sriperumbudur, K. Fukumizu, and G.~R. Lanckriet.
\newblock Universality, characteristic kernels and {RKHS} embedding of
  measures.
\newblock {\em Journal of Machine Learning Research}, 12(70):2389--2410, 2011.

\bibitem{Stuart10}
A.~M. Stuart.
\newblock Inverse problems: a {B}ayesian perspective.
\newblock {\em Acta Numer.}, 19:451--559, 2010.

\bibitem{tihonov1963solution}
A.~N. Tihonov.
\newblock Solution of incorrectly formulated problems and the regularization
  method.
\newblock {\em Soviet Math.}, 4:1035--1038, 1963.

\bibitem{yao2022mean}
R. Yao, X. Chen, and Y. Yang.
\newblock Mean-field nonparametric estimation of interacting particle systems.
\newblock In {\em Conference on Learning Theory}, pages 2242--2275. PMLR, 2022.

\bibitem{you2022_DatadrivenPeridynamic}
H. You, Y. Yu, S. Silling, and M. D’Elia.
\newblock A data-driven peridynamic continuum model for upscaling molecular
  dynamics.
\newblock {\em Computer Methods in Applied Mechanics and Engineering},
  389:114400, 2022.

\bibitem{you2021_DatadrivenLearning}
H. You, Y. Yu, N. Trask, M. Gulian, and M. D'Elia.
\newblock Data-driven learning of nonlocal physics from high-fidelity synthetic
  data.
\newblock {\em Computer Methods in Applied Mechanics and Engineering},
  374:113553, 2021.

\bibitem{yuan2010reproducing}
M. Yuan and T.~T. Cai.
\newblock {A reproducing kernel Hilbert space approach to functional linear
  regression}.
\newblock {\em The Annals of Statistics}, 38(6):3412--3444, 2010.

\bibitem{Zellner1980gprior}
A. Zellner and A. Siow.
\newblock Posterior odds ratios for selected regression hypotheses.
\newblock {\em Trabajos de Estadistica Y de Investigacion Operativa},
  31:585--603, 1980.

\end{thebibliography}
}

\end{document}